\documentclass[pdflatex,sn-basic]{sn-jnl}

\usepackage{graphicx}
\usepackage{amsmath,amssymb,amsfonts}
\usepackage{booktabs}
\usepackage{multirow}
\usepackage{algorithm}
\usepackage{algpseudocode}
\usepackage[title]{appendix}
\usepackage{xcolor}
\usepackage{tikz}
\usepackage{afterpage}
\usepackage{pdflscape}
\usepackage{array}
\usetikzlibrary{arrows.meta,positioning,fit,calc}
\definecolor{stgray}{RGB}{241,239,232}
\definecolor{stteal}{RGB}{225,245,238}
\definecolor{stamber}{RGB}{250,238,218}
\definecolor{stred}{RGB}{252,235,235}
\definecolor{stblue}{RGB}{230,241,251}
\definecolor{bgray}{RGB}{95,94,90}
\definecolor{bteal}{RGB}{15,110,86}
\definecolor{bamber}{RGB}{133,79,11}
\definecolor{bred}{RGB}{163,45,45}
\definecolor{bblue}{RGB}{24,95,165}

\newcommand{\cdvalue}{2.09}
\newcommand{\numpokergap}{$-$0.1}
\newcommand{\numrbdsd}{0.0003}
\newcommand{\numrbdrosd}{0.0184}
\newcommand{\numwinsrhoresolvedall}{0}

\newcommand{\numcabwinsone}{10}

\newcommand{\numcabmonofail}{7}
\newcommand{\numcabaccgradual}{0.254}
\newcommand{\numcabacccover}{0.425}
\newcommand{\numablcommon}{19}
\newcommand{\numablflips}{10}
\newcommand{\numablbothres}{4}

\newcommand{\numablbconf}{9}
\newcommand{\numpfabr}{0}
\newcommand{\numpfrec}{13,030}
\newcommand{\numwtrainb}{$-$43,507}
\newcommand{\numthlabr}{$-$6,512}
\newcommand{\numthlrec}{6,708}
\newcommand{\numpwtotal}{325}
\newcommand{\numpwres}{288}
\newcommand{\numpwfavrar}{138}
\newcommand{\numpwfavcomp}{150}

\providecommand{\numablcommon}{}\renewcommand{\numablcommon}{21}
\providecommand{\numablflips}{}\renewcommand{\numablflips}{10}
\providecommand{\numablbothres}{}\renewcommand{\numablbothres}{4}
\providecommand{\numhthreerarpre}{}\renewcommand{\numhthreerarpre}{60}
\providecommand{\numhthreebinpre}{}\renewcommand{\numhthreebinpre}{510}
\providecommand{\numhthreerarfrac}{}\renewcommand{\numhthreerarfrac}{0.46}
\providecommand{\numhthreebinfrac}{}\renewcommand{\numhthreebinfrac}{0.68}

\newcommand{\fampone}{0.007}
\newcommand{\famphund}{0.306}

\newcommand{\lobototal}{14}
\newcommand{\lobonotsep}{13}
\newcommand{\lobodissolvehund}{3}

\newcommand{\lobopmaxone}{$<$0.001}

\newcommand{\hfourabruptretrains}{5.8}

\newcommand{\hfourabruptrollbacks}{2.2}
\newcommand{\hfourgradualretrains}{6.6}
\newcommand{\hfourgradualalerts}{46.9}
\newcommand{\hfourgradualrollbacks}{2.3}
\newcommand{\hfourrecurrentretrains}{9.2}
\newcommand{\hfourrecurrentalerts}{48.6}
\newcommand{\hfourrecurrentrollbacks}{3.3}
\newcommand{\cyclefirst}{4{,}790}
\newcommand{\cyclecorr}{6{,}390}
\newcommand{\noaacyclescorr}{2.8}

\newcommand{\labelwait}{2{,}690}
\newcommand{\canarypart}{3{,}700}
\newcommand{\feaslatency}{10{,}400}
\newcommand{\powdetsixlow}{0.017}
\newcommand{\powdetsixhigh}{0.034}
\newcommand{\powdetthreelow}{0.024}
\newcommand{\powdetthreehigh}{0.047}
\newcommand{\hfivewin}{1}
\newcommand{\hfivelose}{11}
\newcommand{\hfiveddlower}{6}

\newcommand{\friedmanphund}{0.045}
\newcommand{\friedmanptwohund}{0.077}

\newcommand{\eonefire}{22}

\newcommand{\eonear}{$-$23{,}361}

\newcommand{\eonebin}{$-$25{,}819}

\newcommand{\eonero}{$-$18{,}055}

\newcommand{\eonerecall}{0.005}
\newcommand{\etwowin}{1}
\newcommand{\etwolose}{11}
\newcommand{\apiequiv}{390}
\newcommand{\apiequivtotal}{390}
\newcommand{\ethreeagreeone}{4}
\newcommand{\ethreeagreehund}{3}
\newcommand{\hsixddseeds}{4}
\newcommand{\runtworaw}{2{,}340}

\begin{document}

\title[Retrain, Alert, or Rollback?]{Retrain, Alert, or Rollback? A Three-Response Decision Framework for Production ML Under Concept Drift}

\author[1]{\fnm{Ali} \sur{Sadeghian}}\email{ali.sadeghian@ee.sharif.edu}\equalcont{These authors contributed equally to this work.}

\author[1]{\fnm{Alireza} \sur{Shafaei}}\email{ali.shafaei.80@sharif.edu}\equalcont{These authors contributed equally to this work.}

\author*[2]{\fnm{Hamid} \sur{Hadibeik}}\email{h.beik@alumni.stanford.edu}

\author[1]{\fnm{Babak Hossein} \sur{Khalaj}}\email{khalaj@sharif.edu}

\affil[1]{\orgdiv{Department of Electrical Engineering}, \orgname{Sharif University of Technology}, \orgaddress{\city{Tehran}, \country{Iran}}}

\affil*[2]{\orgname{Independent Researcher}, \orgaddress{\city{San Jose}, \state{CA}, \country{USA}}}

\abstract{When a deployed model degrades under concept drift, an operator can retrain
it, raise an alert and monitor, or roll back. Yet most
drift-adaptation methods consider only retraining. We present RAR
(Retrain--Alert--Rollback), a state machine that gives each response an
explicit trigger and cost. Retraining is gated on drift detection, label
availability, and measured degradation. Every retrained model must pass a two-stage canary before promotion, and corruption-triggered and quality-triggered rollbacks enter different hold states. Under a prospectively
specified protocol we compare RAR with six baselines, including a cost-aware
retraining policy, on 13 streams: eight real public streams and five
semi-synthetic streams derived from them. RAR does not dominate. At low
retraining cost, retrain-heavy baselines lead, and no policy's rank differs significantly from RAR's. At higher cost, RAR beats detector-triggered
retraining, periodic retraining, and no response simultaneously on four streams, each within a bounded cost window. Its advantage over detector-triggered retraining
is consistent with a latency-feasibility condition stated in advance. Against
the cost-aware policy, results are heterogeneous. RAR retrains before true
drift onsets one eighth to one third as often as detector-triggered retraining,
but at lower net utility on most semi-synthetic streams. On H2's designated corruption stream (a registered follow-up), rollbacks fire
yet RAR loses to retraining baselines.
At $\rho = 1$, removing the canary, and thus rollback, raises utility on 11 of 13 streams. On a post-hoc
follow-up stream, cause-specific hold release resumes adaptation where a single cooldown does not. We provide the harness, configurations, and per-cell
results for reproduction.}

\keywords{Concept drift, Model maintenance, Retraining policies, Canary deployment, Data streams, MLOps}

\maketitle

\section{Introduction}
Machine learning models deployed in production degrade. The joint
distribution $P(X, Y)$ that generated the training data shifts, whether
seasonally, abruptly, or through slow environmental change. As it does, the
model's predictions become progressively less reliable, often without any
explicit signal that this is occurring
\citep{gama2014survey,lu2019learning,vela2022temporal}. The literature calls
this phenomenon \emph{concept drift}, and it has produced two mature bodies
of work in response. Drift \emph{detection} asks whether the distribution has
changed
\citep{gama2004learning,bifet2007adwin,page1954continuous,raab2020kswin,ross2012ecdd}.
Drift \emph{adaptation} modifies the model once change is suspected,
typically by retraining or by maintaining adaptive ensembles
\citep{domingos2000mining,gomes2017arf,kolter2007dwm}.

Both bodies of work share an assumption so pervasive that it is rarely
stated: when drift is detected, the response is to update the model. The
decision facing the system is treated as binary, retrain or do not retrain,
and research effort concentrates on making the trigger for that decision more
accurate.

Practitioners who operate production ML systems know that this binary framing
is incomplete, because two additional responses are part of their routine practice. The first is to \emph{withhold action and escalate}. Drift may be
suspected, but labels are too scarce, too delayed, or too unreliable to
retrain safely. The correct response is then to alert a human operator or
intensify monitoring while continuing to serve the current model. The second
is to \emph{revert}. A retrained model, once deployed, sometimes performs
worse than its predecessor, and the correct response is then to roll back to
the previous version. This failure is most acute when the apparent ``drift''
was in fact upstream data corruption, so that retraining fits the model to corrupted data.
Interview studies of ML practitioners document both behaviors as routine
\citep{shankar2022operationalizing}. The software engineering literature has
long institutionalized the second under the name \emph{canary deployment}
\citep{schermann2018continuous,schermann2016bifrost}.

Recent work has begun to treat drift response as a budgeted decision
problem. Work on when to retrain under budget and latency limits includes
RCCDA \citep{piaseczny2025rccda} and two recent preprints (Beytur et al.,
arXiv:2512.12816; Dasari, arXiv:2608.19488). Other work builds controllers
that map certified risk bounds to interventions, including rollback
\citep{lamaakal2026drift2act}. What remains missing is a framework in which
retraining, monitored escalation, and rollback are distinct actions, each with
its own trigger conditions and costs, and that evaluates them together under
label delay against the policies practitioners already use.

This paper addresses that gap. We make three contributions.

\textbf{First, we formalize the three-way decision.} We define RAR
(Retrain--Alert--Rollback), a decision framework in which a policy observes a
drifting data stream and, at each decision point, selects among four actions:
continue, retrain, alert, or roll back. Each action has an explicit cost and
an explicit trigger condition stated in terms of observable quantities
(detector state, label availability, measured degradation of the incumbent,
and canary-window performance). Retraining requires all three gates to pass. Alerting is the designated response when drift evidence
exists but the retraining gates fail. Rollback is triggered by a two-stage
canary evaluation that every retrained model must pass before full
deployment. A two-tier post-rollback locking protocol separates
corruption-triggered from quality-triggered rollbacks. The framework is
expressed as a state machine
(Figure~\ref{fig:statemachine}) whose states and transitions map directly
onto components of standard MLOps infrastructure.

\begin{sloppypar}
\textbf{Second, we formalize the operator's objective as cost-accounted
sequential decision-making} and derive the conditions under which each arm is
admissible. The objective is cumulative prediction utility minus the
accumulated costs of retraining, alerting, and rollback. It extends the
cost-sensitive adaptation question posed by \citet{zliobaite2015cost} (``when
is it worth updating the model?'') to a richer action space. In that space,
\emph{not updating} is itself differentiated into monitored inaction (alert)
and reversal (rollback).
\end{sloppypar}

\textbf{Third, we evaluate three-arm drift-response policies under a
prospectively specified protocol.} \begin{sloppypar}We evaluate RAR against six
baselines on eight real public streams and five semi-synthetic streams with
ground-truth drift onsets. The baselines are no-response, always-retrain,
detector-triggered binary retraining, single-arm policies, and a
pilot-informed cost-aware retraining baseline whose definition was committed
before its full-scale results. The evaluation spans a broad range of
retraining costs and includes analytic break-even curves for the cost-blind
policies. Policies that contain a stochastic canary run with 30 routing
seeds; deterministic policies give identical outputs across seeds. We use
rank-based statistical comparison following \citet{demsar2006statistical}.
The semi-synthetic streams contain injected abrupt, gradual, and recurrent drift; a
feature-corruption episode rather than genuine concept drift; and a
corruption-triggered rollback placed immediately before genuine abrupt
drift. The corruption stream operationalizes the scenario that motivates the
rollback arm, because a binary policy that retrains on corrupted data has no
mechanism for recovery, while RAR does. The rollback-then-drift stream
stress-tests the framework's post-rollback recovery.\end{sloppypar}

Our central hypothesis, specified in advance, was that the three-arm policy
uniformly dominates binary and single-arm policies, that is, it is never
meaningfully worse and is substantially better precisely in the regimes
practitioners fear most: scarce labels, expensive retraining, and corrupted
inputs. That hypothesis is \emph{not} supported in its uniform form. At the
cheap retraining costs of the ratified grid, retrain-heavy baselines lead, and no policy's rank differs significantly from the three-arm policy's. The specified
conditional branch is what the evidence supports, and it is the more useful
result. It comprises stream-specific bounded operating windows of retraining cost inside which graduated response simultaneously beats Binary, Always-Retrain, and
No-Response, and an advantage over Binary whose pattern is consistent with the framework's own latency-feasibility condition. These come together with operational properties that mean net utility does not capture: fewer retrains before true onsets and, on a post-hoc follow-up stream, recovery from corruption-triggered holds.
Section~\ref{sec:results} reports each of six hypotheses under the
prospectively specified protocol. These include H2, which is not supported on
its specified instrument; H3, which holds only in its false-retrain
component; H5, whose policy-level evidence runs against it; and H6, which is
null on its specified stream.

\section{Background and Related Work}
\label{sec:related}
RAR sits at the intersection of five literatures. Table~\ref{tab:positioning}
summarizes the positioning.

\subsection{Drift Detection}
Drift detectors monitor a stream statistic (error rate, feature
distributions, or prediction distributions) and raise a flag when change is
statistically supported. DDM \citep{gama2004learning} monitors the online
error rate against its minimum. ADWIN \citep{bifet2007adwin} maintains an
adaptive window and cuts it when two sub-windows differ significantly.
Page--Hinkley \citep{page1954continuous} applies CUSUM-style sequential change
detection, KSWIN \citep{raab2020kswin} applies Kolmogorov--Smirnov tests over
sliding windows, and ECDD \citep{ross2012ecdd} applies EWMA control charts.
This literature answers \emph{whether} change occurred. It is silent on
\emph{what to do next}. The implicit contract is that a detection event
triggers retraining, and detector quality is evaluated by detection delay and
false-alarm rate rather than by the utility of the downstream response. RAR
consumes any of these detectors as Gate~1 of its retrain arm and treats the
response decision as the object of study.

\subsection{Drift Adaptation and Online Learning}
Adaptive learners modify the model continuously or upon detection. Hoeffding
trees grow incrementally \citep{domingos2000mining}, Adaptive Random Forests
replace drifting ensemble members \citep{gomes2017arf}, Dynamic Weighted
Majority reweights experts \citep{kolter2007dwm}, and DriftSurf alternates
between stable and reactive states \citep{tahmasbi2021driftsurf}. These
methods embed the adaptation decision inside the learner. They do not model
adaptation \emph{cost}, and they have no notion of withholding adaptation
because labels are unavailable. Critically, they have no mechanism to detect
that an adaptation step made the model worse and reverse it. RAR is
learner-agnostic: it wraps any base learner and makes the adaptation decision
explicit, costed, and reversible. CALIPER \citep{fujiwara2026when} is closest
to RAR's Gate~2. It tests from unlabeled data whether enough post-drift data
has accumulated to retrain, adapting the required sample size to the stream.
RAR instead uses a fixed label-availability threshold $\theta_L$ as one
component of a larger action-selection and deployment state machine. An
adaptive sufficiency test such as CALIPER could replace that threshold.

\subsection{Cost-Sensitive Adaptation}
The closest prior work to our objective function is \citet{zliobaite2015cost},
who ask when updating a model is worth its cost and propose a
return-on-investment framing for binary retrain decisions. We adopt their
central insight, that adaptation must be justified against its cost, and
extend it in two directions. The first is from a binary action space to a
four-action space with differentiated costs ($C_r$, $C_a$, $C_b$). The second
is from an analytical framing to a controlled empirical comparison across
drift types. Matchmaker \citep{mallick2022matchmaker} addresses cost-efficient
drift mitigation at scale by matching test points to training batches.
Performance-aware drift detection is surveyed in
\citep{bayram2022degradation}. Closer to our question, RCCDA
\citep{piaseczny2025rccda} derives threshold policies for model updates under
a resource budget. Two recent preprints address the same question: Beytur
et al.\ (2025, arXiv:2512.12816) allocate training and deployment resources
under sudden drift, and Dasari (2026, arXiv:2608.19488) compares retraining
policies empirically under budget and latency constraints. These three
budgeted-retraining methods all concern \emph{when to retrain}, with no alert
or rollback action. Drift2Act
\citep{lamaakal2026drift2act} is the most closely related work. It is a controller that
maps drift evidence and an anytime-valid risk bound to interventions ranging
from recalibration to retraining and rollback under budgets and cooldowns.
RAR differs in what it makes explicit and tests: label delay as two clocks,
canary-gated promotion with typed rollback causes and hold states, and a
prospectively specified comparison against the binary and single-arm
policies in current use.

\subsection{Safe Deployment and Canary Releases}
Staged rollout of new software versions, known as canary deployment, is
standard practice in software engineering \citep{schermann2018continuous}.
Automated multi-phase live-testing strategies have also been studied
academically \citep{schermann2016bifrost}. Controlled online experimentation
at scale \citep{kohavi2013online} provides the statistical foundations for
comparing a new variant against a running baseline on live traffic. This
literature, however, concerns \emph{software} releases evaluated on
engagement or system metrics. Its transfer to \emph{model} releases under
concept drift has received little formal treatment. In that setting the
comparison window itself sits on a shifting distribution, and the ``new
version'' may be corrupted by the very drift it responds to. RAR's
contribution here is to make canary evaluation a typed transition in a
drift-response state machine. RAR treats window sizes and rollback margins
as experimental variables (Section~\ref{sec:ablations}) rather than fixed
operational folklore.

\subsection{MLOps Systems and Practitioner Studies}
Systems papers and practitioner studies document the operational reality that
motivates RAR. They cover hidden technical debt in ML systems
\citep{sculley2015hidden}, deployment challenges across the ML lifecycle
\citep{paleyes2022challenges}, and interview evidence that practitioners
routinely monitor, escalate to humans, and revert bad deployments
\citep{shankar2022operationalizing}. These works establish that the three
responses exist in practice and that automated retraining without safeguards
is operationally dangerous. They stop short of formalizing the decision
problem, stating trigger conditions, or evaluating policies under controlled,
reproducible conditions. This is the gap RAR addresses.

\afterpage{\begin{landscape}
\begin{table}
\caption{Feature matrix positioning RAR against representative prior work.}
\label{tab:positioning}
\footnotesize
\setlength{\tabcolsep}{3pt}
\begin{tabular}{@{}>{\raggedright\arraybackslash}p{6.2cm}*{6}{>{\centering\arraybackslash}p{1.95cm}}@{}}
\toprule
Work & Detects drift & Costs adaptation & Alert arm & Rollback arm & Canary for models & Policy-level eval. \\
\midrule
Detectors \citep{gama2004learning,bifet2007adwin,raab2020kswin,ross2012ecdd} & \checkmark & --- & --- & --- & --- & (of detectors) \\
Adaptive learners \citep{gomes2017arf,kolter2007dwm,tahmasbi2021driftsurf} & (implicit) & --- & --- & --- & --- & (of learners) \\
\citet{zliobaite2015cost} & \checkmark & \checkmark (binary) & --- & --- & --- & partial \\
Matchmaker \citep{mallick2022matchmaker} & \checkmark & \checkmark & --- & --- & --- & (systems eval) \\
Budgeted retraining \citep{piaseczny2025rccda}; Beytur et al.; Dasari & (implicit) & \checkmark (budget) & --- & --- & --- & \checkmark (retrain timing) \\
Drift2Act \citep{lamaakal2026drift2act} & \checkmark & \checkmark (budget) & (escalation) & \checkmark & --- & \checkmark \\
Canary / experimentation \citep{schermann2018continuous,schermann2016bifrost,kohavi2013online} & --- & --- & --- & \checkmark (software) & \checkmark (software) & --- \\
Practitioner studies \citep{sculley2015hidden,paleyes2022challenges,shankar2022operationalizing} & (descriptive) & (descriptive) & (descriptive) & (descriptive) & (descriptive) & --- \\
\textbf{RAR (this paper)} & \checkmark (pluggable) & \checkmark (3 costs) & \checkmark & \checkmark & \checkmark (models) & \checkmark \\
\bottomrule
\end{tabular}
\end{table}
\end{landscape}}

\section{Problem Formulation}
\label{sec:formulation}
Notation introduced in Sections~\ref{sec:formulation}--\ref{sec:design} is
collected, with defaults, in the glossary of Appendix~\ref{app:glossary}.

\subsection{Stream Model and Prequential Evaluation}
Let $S = ((x_t, y_t))_{t=1}^{T}$ be a data stream with features $x_t \in
\mathcal{X}$ and labels $y_t \in \mathcal{Y}$, generated by a time-varying
joint distribution $P_t(X, Y)$. Following \citet{gama2014survey}, a change in $P_t(X, Y)$ decomposes into
\emph{real concept drift}, a change in $P_t(Y \mid X)$; \emph{virtual} or
covariate drift, a change in $P_t(X)$ with $P(Y \mid X)$ fixed; and label
shift, a change in $P_t(Y)$. We use ``drift'' for real concept drift and
distinguish \emph{abrupt}, \emph{gradual}, \emph{incremental}, and
\emph{recurrent} forms. \emph{Data corruption} is an upstream fault that
alters the observed features. Formally it is a change in the observed
$P_t(X)$, but not a change in the relation the model should learn, so
adapting to it is an error rather than a repair. Corruption and real drift
can look alike to an error-rate detector, and this is what motivates the
rollback arm.

Evaluation is prequential (test-then-train) \citep{gama2014survey}: at each
step $t$ the deployed model produces $\hat{y}_t$ before $y_t$ is revealed.
Labels arrive with delay $\tau \geq 0$. The \emph{label availability} in a
window $W$ at time $t$ is
\begin{equation}
\ell_t(W) = \frac{1}{|W|} \sum_{i \in W} \mathbb{1}[y_i \text{ observed by
time } t].
\label{eq:labelavail}
\end{equation}
Label delay partitions all of the framework's bookkeeping into two clocks.
Quantities that require labels (rolling accuracies, $\ell_t$, and the
canary comparison) are computed at the \emph{label frontier} $t - \tau$
and can never read past it. The trailing $\tau$ instances of any such
window are unlabeled by construction. Quantities computed on features alone
(the quality statistic $q_t$ of Section~\ref{sec:rollbackarm}) or on
elapsed time (hold cooldowns and canary timeouts) run at the \emph{live
edge} $t$ with no delay. This includes the corruption condition R2, which
therefore reacts to a degenerate feed without label latency. The
distinction is what allows $q_t$ to certify recovery during a hold while
every accuracy-based signal is still blind. The released harness enforces
this distinction as a property test: no frontier-clocked computation may read an instance
newer than $t - \tau$, and no live-edge mechanism may block on a label.

\subsection{Policies and Actions}
A \emph{drift-response policy} $\pi$ maps the observable history $H_t$
(predictions, revealed labels, detector state, model version history) to an
action
\begin{equation}
a_t = \pi(H_t) \in \{\textsc{Continue}, \textsc{Retrain}, \textsc{Alert},
\textsc{Rollback}\}.
\label{eq:actions}
\end{equation}
\textsc{Retrain} fits a new model $M_{\text{new}}$ on a trailing window
$W_{\text{train}}$ and initiates staged deployment. \textsc{Alert} emits a
logged escalation event and continues serving the incumbent model
$M_{\text{old}}$. It differs from \textsc{Continue} in its explicit trigger
condition, its cost, and its side effect of starting a secondary monitoring
clock. \textsc{Rollback} discards $M_{\text{new}}$, restores
$M_{\text{old}}$, resets detector state, and enters a mandatory monitored
hold state. During the hold, \textsc{Retrain} is inadmissible except under
the emergency-override conditions of Section~\ref{sec:twotier}.

\subsection{Objective}
Each action has a cost: $C_e$ per prediction error (normalized to $C_e = 1$),
$C_r$ per retrain, $C_a$ per alert, and $C_b$ per rollback. Writing
$N_r(\pi)$, $N_a(\pi)$, $N_b(\pi)$ for the number of times each arm fires
over the stream, the \emph{net utility} of policy $\pi$ on stream $S$ is
\begin{equation}
U(\pi; S) = \sum_{t=1}^{T} \mathbb{1}[\hat{y}_t = y_t]
  - C_r N_r(\pi) - C_a N_a(\pi) - C_b N_b(\pi).
\label{eq:netutility}
\end{equation}
Equation~(\ref{eq:netutility}) embeds the error cost implicitly rather than
as an explicit penalty term. Since the utility mapping is binary and the
stream length $T$ is fixed per configuration, $U = T - C_e N_{\text{err}} -
C_r N_r - C_a N_a - C_b N_b$ with $C_e = 1$. Maximizing $U$ is therefore
exactly equivalent to minimizing total incurred cost, and $\rho = C_r / C_e$
is well-defined. The binary mapping generalizes directly to asymmetric error
costs, with false positives priced differently from false negatives, as in
fraud decisioning. The released harness supports this generalization,
though our experiments use the symmetric case.

We parameterize cost regimes by the \emph{cost ratio} $\rho = C_r / C_e$.
As defaults we fix $C_a = 0.05\,C_r$ and $C_b = 0.2\,C_r$, reflecting that
alerts consume human review time and rollbacks consume deployment overhead
but no training compute; both ratios are ablated. The operator's problem is
to select $\pi$ maximizing $\mathbb{E}[U(\pi; S)]$ over the drift process.
We compare policies by their paired differences in net utility in each
configuration and by their ranks within each stream.

Three remarks on design trade-offs. First, we deliberately assign
\textsc{Alert} a nonzero cost. If alerting carried no operational cost and
did not alter subsequent decisions, the utility ledger would be indifferent
between \textsc{Alert} and continuing. A positive $C_a$ prevents unlimited
escalation from being costless, and the question becomes whether
\emph{costed} graduated response pays. Second, we charge
\textsc{Rollback} less than \textsc{Retrain} because reverting to a cached
model version requires rollout overhead but no training. If an
organization's registry makes rollback nearly free, our $C_b$ errs against RAR, which makes any positive finding conservative. Third, every RAR decision rule is deliberately \emph{cost-blind}. The cost
model is read by the utility ledger and by nothing else, a property
asserted by a test in the released harness. Net utility is therefore
exactly affine in $\rho$ for every cost-blind policy, which
Section~\ref{sec:design} exploits. Cost-\emph{sensitivity} enters at the
level of policy selection: the reported break-even curves tell an operator,
for their own $\rho$, which policy to run. Making a gate itself read $\rho$
is a natural extension that we specify for one baseline
(Section~\ref{sec:policies}) and otherwise leave to future work.

\section{The RAR Framework}
\label{sec:framework}
\begin{figure}[t]
  \centering
  \begin{tikzpicture}[
    font=\footnotesize,
    st/.style={draw, rounded corners=2.5pt, align=center, inner sep=3.5pt,
               minimum width=3.1cm, minimum height=0.95cm, line width=0.5pt},
    lab/.style={font=\scriptsize, inner sep=1.5pt, align=center},
    e/.style={-{Stealth[length=2.2mm]}, line width=0.6pt}
  ]
    \definecolor{stgray}{RGB}{241,239,232}
    \definecolor{stteal}{RGB}{225,245,238}
    \definecolor{stamber}{RGB}{250,238,218}
    \definecolor{stred}{RGB}{252,235,235}
    \definecolor{bgray}{RGB}{95,94,90}
    \definecolor{bteal}{RGB}{15,110,86}
    \definecolor{bamber}{RGB}{133,79,11}
    \definecolor{bred}{RGB}{163,45,45}

    \node[st, fill=stgray, draw=bgray] (srv) at (0, 0)
      {\textsc{Serving}\\ \scriptsize incumbent live};
    \node[st, fill=stteal, draw=bteal] (c1) at (-2.65, -2.95)
      {\textsc{Canary-1}\\ \scriptsize $p_1$ routed to $M_{\text{new}}$};
    \node[st, fill=stteal, draw=bteal] (c2) at (2.65, -2.95)
      {\textsc{Canary-2}\\ \scriptsize $M_{\text{new}}$ serves all};
    \node[draw=bteal, dashed, rounded corners=4pt, inner sep=7pt,
          fit=(c1)(c2)] (can) {};
    \node[lab, anchor=north, text=bteal] at (can.south) {two-stage canary};
    \node[st, fill=stamber, draw=bamber, dashed] (hs) at (-2.65, -5.55)
      {\textsc{Alert-Hold-Soft}\\ \scriptsize quality rollback};
    \node[st, fill=stred, draw=bred, line width=1.1pt] (hh) at (2.65, -5.55)
      {\textsc{Alert-Hold-Hard}\\ \scriptsize corruption rollback};

    \draw[e, draw=bgray] (srv.west) -| node[lab, pos=0.75, left]
      {retrain:\\ G1--G3 pass} (c1.north);
    \draw[e, draw=bteal] (c1.east) -- node[lab, above] {Stage~1\\ passed} (c2.west);
    \draw[e, draw=bteal] (c2.north) |- node[lab, pos=0.25, right]
      {commit} (srv.east);

    \draw[e, draw=bgray] (srv.south) ++(-0.35,0)
      .. controls +(-0.25,-0.9) and +(0.25,-0.9) .. ++(0.7,0);
    \node[lab, anchor=north] at (0,-1.2) {alert (A1--A3):\\ stay in \textsc{Serving}};

    \draw[e, draw=bamber] (can.south -| hs.north) --
      node[lab, right, pos=0.45] {R1, R3\\ (either stage)} (hs.north);
    \draw[e, draw=bred] (can.south -| hh.north) --
      node[lab, left, pos=0.45] {R2\\ (either stage)} (hh.north);

    \draw[e, draw=bamber] (hs.west) -- ++(-0.55,0) |- ($(srv.north west)+(0.5,0.45)$)
      -- ($(srv.north west)+(0.5,0)$);
    \node[lab, anchor=south east] at ($(srv.north west)+(0.35,0.47)$)
      {$\kappa_{\text{short}}$ expiry or override};
    \draw[e, draw=bred] (hh.east) -- ++(0.55,0) |- ($(srv.north east)+(-0.5,0.45)$)
      -- ($(srv.north east)+(-0.5,0)$);
    \node[lab, anchor=south west] at ($(srv.north east)+(-0.35,0.47)$)
      {$\kappa_{\text{long}}$ expiry or $q_t$ clean (early exit)};
  \end{tikzpicture}
  \caption{The RAR state machine. Retraining is admissible only from
  \textsc{Serving} when all three gates pass; every retrained model traverses
  the two-stage canary; rollbacks are typed by cause (R1 canary failure, R2
  data corruption, R3 sanity failure) and may fire from either
  canary stage, entering a hard hold (corruption-triggered; thick border; no
  override) or a soft hold (quality-triggered; dashed border) with the
  severity-gated emergency override of Section~\ref{sec:twotier}.}
  \label{fig:statemachine}
\end{figure}
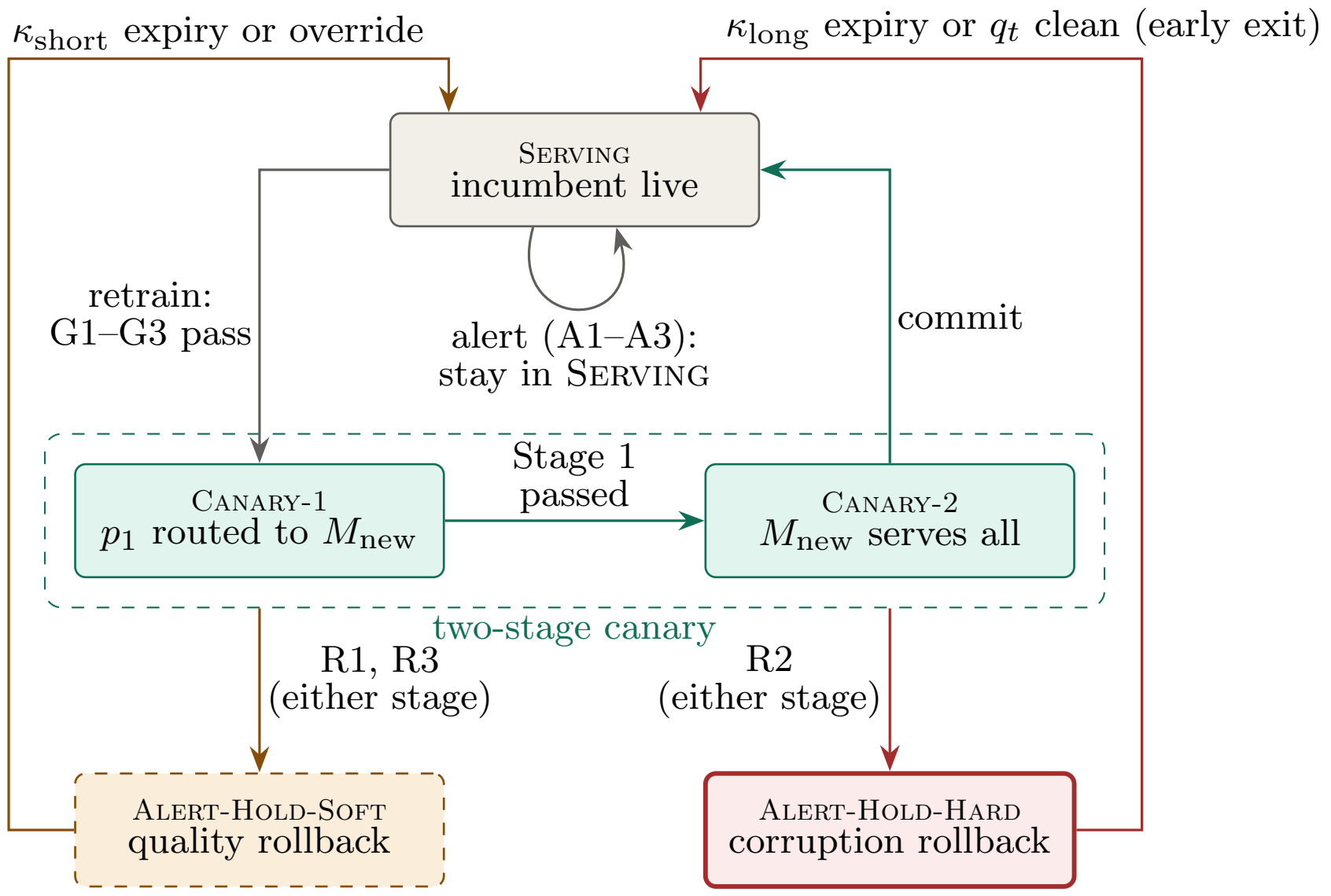

RAR is a state machine over five states---\textsc{Serving},
\textsc{Canary-1}, \textsc{Canary-2}, \textsc{Alert-Hold-Soft}, and
\textsc{Alert-Hold-Hard}---with transitions governed by three gated arms and
an upstream corruption pre-filter.
Figure~\ref{fig:statemachine} shows the machine. \textsc{Serving} is the
steady state. A retrain event moves the system through the two canary
states, back to \textsc{Serving} on commit or to one of the two hold states
on rollback; the hold is classified by rollback cause
(Section~\ref{sec:twotier}).
Quality-triggered holds admit a severity-gated emergency override, while
corruption-triggered holds persist until the corruption clears or
$\kappa_{\text{long}}$ expires.

\subsection{Retrain Arm}
\textsc{Retrain} is admissible only when three gates pass jointly at time
$t$:
\begin{itemize}
\item \textbf{Gate 1 --- Drift confirmed.} A pluggable detector $D$ (ADWIN,
DDM, KSWIN, Page--Hinkley, or ECDD in our experiments) has raised a change
flag at estimated onset $\hat{t}_d$. Detection is suppressed for a short
burn-in after each model birth ($g$ plus the minimum reference-window size,
in frontier instances), and the detector re-baselines at its end. Without this
burn-in, a detector fed a freshly frozen or committed model's errors flags the
transient itself, and no admissible reference window exists that early.
Detector
warm-up periods are likewise sized so that a reset detector must accumulate
evidence comparable to $W_{\text{cur}}$ before it may fire again.
\item \textbf{Gate 2 --- Labels available.} $\ell_t(W_{\text{train}}) \geq
\theta_L$, with default $\theta_L = 0.7$. Retraining on a sparsely labeled
window gives a high-variance fit from too few labeled examples. This gate
operationalizes the practitioner heuristic of waiting for labels. The
retraining window is anchored at the estimated onset $\hat{t}_d$ and
extends forward to the most recent labeled instance $t - \tau$, so every
training instance is both post-drift and labeled. The trailing $\tau$
instances of any window are unlabeled by construction. The window grows
only up to its nominal span. Once it reaches $W_{\text{train}}$ it is
fixed, so a Gate-3-delayed retrain trains on the labeled subset of the
first $W_{\text{train}}$ post-onset instances. The cap keeps training-set
size comparable across configurations and policies, and it bounds exposure
to concept reversal under recurrent drift. Consequently the earliest
admissible retrain time is
$t^{*} = \hat{t}_d + \tau + \lceil \theta_L W_{\text{train}} \rceil - 1$,
and feasibility requires $W_{\text{train}} \geq \tau / (1 - \theta_L)$.
Figure~\ref{fig:windows} (Appendix~\ref{app:glossary}) visualizes all
windows on a worked example at the frozen defaults ($\hat{t}_d = 9{,}500$,
$\tau = 800$, $g = 200$, $W_{\text{train}} = 2{,}700$).
\item \textbf{Gate 3 --- Degradation.} Define the \emph{degradation
estimate}
\begin{equation}
\hat{\Delta}_t = \hat{A}_{\text{pre}}(t) - \hat{A}_{\text{cur}}(t),
\label{eq:benefit}
\end{equation}
where $\hat{A}_{\text{pre}}$ is the incumbent model's rolling accuracy over a
reference window ending at $\hat{t}_d - g$, and $\hat{A}_{\text{cur}}$ is
its rolling accuracy over the most recent window $W_{\text{cur}}$. The
\emph{guard gap} $g$ compensates for detection latency. Detectors accumulate
statistical significance over drifted instances before firing, so the window
immediately preceding the trigger is itself contaminated with post-drift
data. Without the guard, $\hat{A}_{\text{pre}}$ is biased downward and
Gate~3 becomes systematically harder to pass. This under-retraining bias is
most severe under gradual drift, where detection latency is longest. We
ablate $g$ in Section~\ref{sec:ablations}. On semi-synthetic streams whose
true onsets are known, we verify that the guard reduces the contamination
bias. Detection lag varies across streams by orders of magnitude while $g$
is fixed, so a residual bias remains where lag greatly exceeds $g$; we
report it rather than claim full removal. Gate 3 passes
iff $\hat{\Delta}_t \geq \theta_R > 0$, that is, when the incumbent has
lost at least $\theta_R$ accuracy relative to its pre-drift reference accuracy.
$\hat{\Delta}_t$ measures loss, not the gain a new model would deliver. It
bounds what retraining can recover only if a model fit on post-onset data
regains the pre-drift accuracy. The gate does not check this condition; the
canary then tests it on live traffic. Gate 3 is therefore a severity threshold, not a benefit
estimate, and it does not read $C_r$: all RAR decision rules are cost-blind
by design (Section~\ref{sec:formulation}). The one policy that prices
$\hat{\Delta}_t$ against $\rho$ is the Cost-Aware Binary baseline
(Section~\ref{sec:policies}).
\end{itemize}
When all three gates pass, the system fits $M_{\text{new}}$ on $W_{\text{train}}$, charges $C_r$, and enters \textsc{Canary-1}.

\subsection{Alert Arm}
\textsc{Alert} fires when \textsc{Retrain} is inadmissible but evidence of
trouble exists---i.e., when any of the following holds:
\begin{itemize}
\item \textbf{A1.} Gate 1 passed but Gate 2 failed: drift is detected but
labels are too scarce to retrain safely.
\item \textbf{A2.} Gates 1--2 passed but Gate 3 failed: drift is detected but
the incumbent has not degraded past $\theta_R$.
\item \textbf{A3.} No detector flag, but a soft anomaly condition holds:
rolling accuracy has fallen more than a soft threshold below its reference,
or the feature-quality statistic (Section~\ref{sec:rollbackarm}) is marginal.
\end{itemize}
\textsc{Alert} charges $C_a$ and logs an escalation event. It also starts
(or refreshes) a secondary monitoring clock that tightens the anomaly
condition A3 for the next $W_{\text{mon}}$ instances.
\textsc{Alert} can \emph{escalate into} \textsc{Retrain}: if labels subsequently accumulate
($\ell_t$ crosses $\theta_L$) while the detector flag is still active, the
retrain arm fires on the next decision cycle. This graduated path---suspect,
monitor, then act when evidence suffices---is precisely what binary policies
cannot express.

\subsection{Rollback Arm and the Two-Stage Canary}
\label{sec:rollbackarm}
Every retrained model must pass a two-stage canary before full deployment.

\textbf{Stage 1 (\textsc{Canary-1}).} A fraction $p_1$ of incoming instances
(default $p_1 = 0.2$) is routed to $M_{\text{new}}$. Both models'
accuracies are tracked on the routed sample until $W_1$ \emph{labeled}
routed instances have accumulated. Under label delay $\tau$, counting
canary windows in labeled instances extends each stage's wall-clock duration by approximately $\tau$. Stage-1 exposure stays bounded at $p_1$.
A hard timeout $\kappa_{\text{canary}}$ bounds the wait. If the label
pipeline stalls past it, the stage \emph{fails closed}---rollback plus
alert---rather than committing an unverified model. During label droughts
we additionally monitor the two models' predictions on routed traffic. The simplest label-free tripwire, rolling back on high disagreement alone, is harmful under drift that inverts the concept. After a label flip,
a \emph{correct} challenger disagrees with the stale incumbent on nearly
every instance, and a disagreement threshold would destroy exactly the
challengers the framework exists to promote. What identifies a broken
challenger without labels is \emph{prediction collapse}: $M_{\text{new}}$
emitting a single class across the window while $M_{\text{old}}$'s
predictions remain varied on the same instances. The monitor therefore
requires collapse, high disagreement, and a varied incumbent jointly before
aborting a canary.

In simulation, both models score every routed instance in parallel, a
reproducible proxy for production traffic splitting. Stage~1 routes a
fraction $p_1$ of traffic by stratified interleaving. In each block of
$1/p_1$ consecutive instances exactly one goes to $M_{\text{new}}$. Once per canary episode, a generator seeded with the run seed plus the canary's index selects that slot. The seed changes nothing else in a run; i.i.d.\
Bernoulli routing is kept as an ablation cell. The utility ledger records
the prediction that was \emph{served}. If the router assigns instance $t$
to $M_{\text{new}}$, then $M_{\text{new}}$'s prediction, including its
errors, enters Eq.~(\ref{eq:netutility}). Canary exposure thus carries its
true accounted cost, and H5 is not biased in RAR's favor. Routing decisions
depend only on the seed and the instance index and are fixed before $y_t$ is
observed, which precludes look-ahead bias. Shadow predictions are never
ledgered.

\textbf{Stage 2 (\textsc{Canary-2}).} $M_{\text{new}}$ receives all traffic
for $W_2$ labeled instances while $M_{\text{old}}$ remains cached and
shadow-scored (comparison only, never ledgered).

\textbf{Corruption pre-filtering.} The feature-quality statistic $q_t$
operates \emph{upstream} of the drift detectors. Instances flagged as
hard-corrupt are quarantined from detector updates, which keeps corrupted
data out of detector sliding windows; the model still serves these instances, and they
are still ledgered. Without this layer, corruption poisons detector memory
and produces alarm churn that persists after rollback, because a post-hoc
detector reset is immediately re-poisoned by ongoing corruption. Three
design rules govern the checks, each forced by a failure mode we observed
on real streams. \emph{(i) Per-feature freeze tolerance:} a feature counts
as frozen only when its constant run exceeds a multiple of its longest
constant run on the clean warmup prefix. A single global run-length
threshold misreads legitimately idle features (a categorical feature sorted by time, a feature with seasonal runs of zeros) as stuck feeds. Under the per-feature rule, a feature
constant throughout warmup is exempt as a special case. \emph{(ii)
Categorical exemption:} low-cardinality columns are excluded from the freeze and variance checks, which are meant for continuous-valued features.
\emph{(iii) A step-size check:} independent resampling of a near-uniform
feature leaves its marginal mean and variance essentially unchanged, so
marginal monitoring is \emph{structurally blind} to it. What resampling
always destroys is temporal structure, so $q_t$ additionally tests the mean
absolute successive difference of each monitored feature against its clean
baseline. The pre-filter catches blatant corruption; subtle corruption that
evades it is meant to be caught downstream by the canary. The two
corruption modes of Section~\ref{sec:datasets} were built to test the two
layers. The base matrix ran the blatant mode, and a registered follow-up
ran the subtle one (Section~\ref{sec:h2results}).

\textsc{Rollback} fires from either canary state if any of the following
holds:
\begin{itemize}
\item \textbf{R1 --- Canary failure.} $\hat{A}_{\text{new}} <
\hat{A}_{\text{old}} - \delta$ over the current canary window, with margin
$\delta$. R1 is an operational screen, not a significance test. The paired
power analysis of Appendix~\ref{app:power} covers the commit bar, not R1;
$\delta$ and the window sizes are ablated in Section~\ref{sec:ablations}.
\item \textbf{R2 --- Data corruption.} A feature-quality statistic $q_t$
flags the incoming window as degenerate: any monitored feature's mean or
variance departs more than $3\sigma$ from its historical baseline, the
null fraction exceeds a threshold, or a feature freezes at a constant value.
Baseline statistics are estimated on warmup data and updated only from
instances that pass the checks, so the baseline cannot drift toward the
corruption it exists to detect. R2 asserts a \emph{sustained} state, not an
event. It fires only after a run of consecutive assessments that indicate corruption, so a single flagged window cannot destroy a challenger. R2 is
evaluated at the live edge (Section~\ref{sec:formulation}). Under
corruption, \emph{both} retraining and continued canary evaluation are
unsafe, since accuracy measured on degenerate inputs certifies nothing.
Reversion to the last known-good model is the conservative action.
\item \textbf{R3 --- Sanity failure.} $M_{\text{new}}$ underperforms a
no-skill baseline (majority-class predictor) on the canary window. R3
complements R1. The canary comparison is \emph{relative} and passes
whenever the challenger beats the incumbent, even when drift has degraded
both. R3 is the \emph{absolute} floor that catches a structurally broken
challenger (inverted labels, wrong feature scale) regardless of the
incumbent's state. R1 is also evaluated at \emph{interim} checkpoints, once a
minimum number of labeled canary instances has accumulated, which bounds
how long a losing challenger can serve. R1 is a margin rule on point
estimates, not a significance test, and repeated interim looks raise its
false-rollback rate above that of a single look. We treat it as an
operational screen and do not claim a calibrated error rate for it. A
sequential test (an alpha-spending boundary or a confidence sequence) is the
calibrated alternative. R3 is deliberately evaluated only on the completed
stage window. Its no-skill reference $\max_c \hat{p}(c)$ is biased upward
at small $n$, and early evaluation rejects sound challengers on sampling
noise. (Algorithm~\ref{alg:rar} shows
both the interim checks and the stage-boundary tests.)
\end{itemize}
\textsc{Rollback} charges $C_b$, discards $M_{\text{new}}$, restores
$M_{\text{old}}$, and resets $D$ (whose statistics are contaminated by the
aborted deployment). Passing both stages is necessary but not sufficient
for promotion. A commit hands the stream, until the next adaptation, to a model judged on one $W_2$ window. Promotion therefore additionally requires
the accuracy difference to clear a one-sided \emph{promotion evidence
threshold} computed on \emph{discordant} pairs only. These are the instances on which exactly one of the two models is correct, the only ones carrying information about which is better. Under the null the discordant wins are
$\mathrm{Binomial}(n_{01}{+}n_{10}, \tfrac{1}{2})$, giving
$z = (n_{01} - n_{10})/\sqrt{n_{01} + n_{10}}$ (McNemar's construction).
Promotion requires $z \geq z_{\text{commit}}$ (default 1, a one-sided
$\alpha \approx 0.16$, so a threshold on evidence rather than a conventional
significance test). The threshold is evaluated once at the end of Stage 2, so
that it is a single-look test. The absence of discordant pairs counts as observed equivalence. The dead band between the rollback margin and the commit bar
reverts to the incumbent, which is the cheap, safe default. Setting
$z_{\text{commit}} = 0$ recovers point-estimate promotion for the ablation.
On commit, $M_{\text{new}}$ becomes the incumbent and the system returns to
\textsc{Serving}.

\subsection{Two-Tier Post-Rollback Locking and Emergency Override}
\label{sec:twotier}
The aftermath of a rollback is \emph{typed} by its cause.
Corruption-triggered rollbacks (R2) enter \textsc{Alert-Hold-Hard} for
$\kappa_{\text{long}}$ instances. During this hold, retraining is
inadmissible with no override---no retrain is safe while inputs are
degenerate. The hold exits early once $q_t$ has been clean for
$m_{\text{clear}}$ consecutive instances (default 200). This hysteresis
prevents a flapping upstream feed from bouncing the system in and out of
the hold.
Quality-triggered rollbacks (R1, R3) enter \textsc{Alert-Hold-Soft} for
$\kappa_{\text{short}}$ instances. During this hold, an \emph{emergency
override} re-admits retraining if all of the following hold simultaneously: a minimum
evidence window $\kappa_{\min}$ has elapsed since the rollback; rolling
accuracy has fallen below the emergency threshold, $\hat{A}_{\text{cur}} <
\hat{A}_{\text{ref}} - \Delta_{\text{crit}}$ with $\Delta_{\text{crit}} \gg
\theta_R$; labels are available, $\ell_t(W_{\text{train}}) \geq \theta_L$;
and $q_t$ is clean.

The two-tier design is meant to avoid a deadlock inherent to any single cooldown. A
uniform lock long enough to prevent rollback--retrain oscillation on
persistent corruption would leave the system blind to genuine severe drift arriving soon after a corruption-triggered rollback (hypothesis H6,
Section~\ref{sec:design}). Requiring $\Delta_{\text{crit}} \gg \theta_R$
prevents the override from immediately re-triggering the retrain that just
failed canary, and $\kappa_{\min}$ prevents override-driven oscillation.
Alerts remain active throughout both hold states. We examine residual
oscillation risk directly in Section~\ref{sec:erroranalysis}.

\subsection{Algorithm}
Algorithm~\ref{alg:rar} presents the full RAR decision loop under prequential
evaluation, and Algorithm~\ref{alg:holds} gives the typed rollback
procedure and hold-state transitions. Gate 3 computes
$\hat{\Delta}_t$ per Eq.~(\ref{eq:benefit}) over
$(W_{\text{ref}}, W_{\text{cur}})$ anchored at the detector's estimated
onset $\hat{t}_d$. Both windows exclude canary
periods to avoid contaminating the incumbent's reference accuracy with
mixed-model predictions.

\begin{algorithm}[tp]
\caption{RAR decision loop (prequential)}
\label{alg:rar}
\footnotesize
\begin{algorithmic}[1]
\Require stream $S$; learner class $\mathcal{F}$; detector $D$; thresholds
$\theta_L, \theta_R, \delta, \Delta_{\text{crit}}$; windows
$W_{\text{train}}, W_{\text{ref}}, W_{\text{cur}}, W_1, W_2,
W_{\text{mon}}$; guard gap $g$; fraction $p_1$; cooldowns
$\kappa_{\text{short}}, \kappa_{\text{long}}, \kappa_{\min},
\kappa_{\text{canary}}$; costs $C_r, C_a, C_b$; seed $s$
\State $M_{\text{old}} \gets$ fit $\mathcal{F}$ on warmup prefix;
\ \textit{state} $\gets$ \textsc{Serving}; \ $c \gets 0$
\For{$t = 1 \dots T$} \Comment{decisions below use frontier statistics and $q_t$}
  \State $\hat{y}_t \gets \textsc{Predict}(\textit{state}, x_t)$
  \Comment{\textsc{Canary-1}: $M_{\text{new}}$ on its seeded slot in each block of $1/p_1$}
  \State compute $q_t$ on $x_t$; store $(t, x_t, \hat{y}_t, q_t)$ in the
  pending-label queue \Comment{live edge}
  \State advance live clocks (holds, canary timeouts, R2 on $q_t$)
  \If{instance $i = t - \tau$ is pending} \Comment{label frontier}
    \State reveal $y_i$; update $\hat{A}_{\cdot}$, $\ell$ with instance $i$
    \State update $D$ with instance $i$ \textbf{only if not} the stored
    $q_i$.corrupt \Comment{quarantine}
    \State \textit{(evaluation only)} ledger $\mathrel{+}=
    \mathbb{1}[\hat{y}_i = y_i]$ for the stored served prediction
  \EndIf
  \If{\textit{state} $=$ \textsc{Serving}}
    \If{$D$.flag}
      \If{$\ell_t(W_{\text{train}}) \geq \theta_L$ \textbf{and}
      $\hat{\Delta}_t \geq \theta_R$}
        \State $M_{\text{new}} \gets$ fit $\mathcal{F}$ on
        $W_{\text{train}}$; ledger $\mathrel{-}= C_r$
        \State \textit{state} $\gets$ \textsc{Canary-1}
      \Else \ \Call{Alert}{\,} \Comment{A1 or A2}
      \EndIf
    \ElsIf{\textsc{Anomaly}($\hat{A}_{\text{cur}}, q_t$)}
    \Call{Alert}{\,} \Comment{A3}
    \EndIf
  \ElsIf{\textit{state} $\in$ \{\textsc{Canary-1}, \textsc{Canary-2}\}
  \textbf{and} (interim R1 fails \textbf{or} \textsc{Collapse}($M_{\text{new}}$)
  \textbf{or} $q_t$.corrupt)}
    \State \Call{Rollback}{R1, collapse, or R2}
    \Comment{interim checks, Sec.~\ref{sec:rollbackarm}}
  \ElsIf{\textit{state} $=$ \textsc{Canary-1} \textbf{and} ($W_1$ labeled
  routed obs.\ \textbf{or} timeout $\kappa_{\text{canary}}$)}
    \If{no timeout \textbf{and} $\hat{A}_{\text{new}} \geq
    \hat{A}_{\text{old}} - \delta$ \textbf{and}
    \textsc{Sane}($M_{\text{new}}$) \textbf{and not} $q_t$.corrupt}
      \State \textit{state} $\gets$ \textsc{Canary-2}
    \Else \ \Call{Rollback}{failed condition}
    \EndIf
  \ElsIf{\textit{state} $=$ \textsc{Canary-2} \textbf{and} ($W_2$ labeled
  obs.\ \textbf{or} timeout $\kappa_{\text{canary}}$)}
    \If{no timeout \textbf{and} $\hat{A}_{\text{new}} \geq
    \hat{A}_{\text{old}} - \delta$ \textbf{and} \textsc{Sane}($M_{\text{new}}$)
    \textbf{and not} $q_t$.corrupt}
      \If{$z \geq z_{\text{commit}}$} \Comment{evidence threshold, one look}
        \State $M_{\text{old}} \gets M_{\text{new}}$; \textit{state} $\gets$
        \textsc{Serving} \Comment{commit}
      \Else \ discard $M_{\text{new}}$; \textit{state} $\gets$
        \textsc{Serving} \Comment{dead band: keep incumbent}
      \EndIf
    \Else \ \Call{Rollback}{failed condition}
    \EndIf
  \ElsIf{\textit{state} $\in$ \{\textsc{Alert-Hold-Hard},
  \textsc{Alert-Hold-Soft}\}}
    \State handle per Algorithm~\ref{alg:holds}
  \EndIf
\EndFor
\Procedure{Alert}{\,} ledger $\mathrel{-}= C_a$; log event; refresh
monitoring clock ($W_{\text{mon}}$); if $\ell_t$ later crosses $\theta_L$
while $D$.flag holds, the next \textsc{Serving} step re-tests the gates
(escalation) \EndProcedure
\end{algorithmic}
\end{algorithm}

\begin{algorithm}[t]
\caption{Typed rollback and hold-state transitions}
\label{alg:holds}
\begin{algorithmic}[1]
\Procedure{Rollback}{\textit{cause}} ledger $\mathrel{-}= C_b$; discard
$M_{\text{new}}$; restore $M_{\text{old}}$; reset $D$; $t_{\text{rb}} \gets
t$
\If{\textit{cause} $=$ R2} \textit{state} $\gets$
\textsc{Alert-Hold-Hard}; $c \gets \kappa_{\text{long}}$; $r \gets 0$
\Else \ \textit{state} $\gets$ \textsc{Alert-Hold-Soft}; $c \gets
\kappa_{\text{short}}$
\EndIf
\EndProcedure
\Statex \textbf{At each step $t$ while holding:}
\If{\textit{state} $=$ \textsc{Alert-Hold-Hard}}
  \State $c \gets c - 1$; \ $r \gets 0$ \textbf{if} $q_t$.corrupt
  \textbf{else} $r + 1$ \Comment{clean run}
  \If{$c = 0$ \textbf{or} $r \geq m_{\text{clear}}$} \textit{state} $\gets$
  \textsc{Serving} \Comment{expiry or corruption cleared}
  \EndIf
\ElsIf{\textit{state} $=$ \textsc{Alert-Hold-Soft}}
  \State $c \gets c - 1$
  \If{$c = 0$} \textit{state} $\gets$ \textsc{Serving}
  \ElsIf{$t - t_{\text{rb}} \geq \kappa_{\min}$ \textbf{and}
  $\hat{A}_{\text{cur}} < \hat{A}_{\text{ref}} - \Delta_{\text{crit}}$
  \textbf{and} $\ell_t(W_{\text{train}}) \geq \theta_L$ \textbf{and not}
  $q_t$.corrupt}
    \State \textit{state} $\gets$ \textsc{Serving}
    \Comment{emergency override: retrain re-admitted}
  \EndIf
\EndIf
\end{algorithmic}
\end{algorithm}

\subsection{Feasibility Conditions and the Gate-2 Sample Penalty}
Two structural constraints bound where the framework can pay. We state them
as prospectively specified conditions rather than discover them post hoc.
First, a \emph{latency feasibility condition}: one full adaptation cycle
--- waiting out the label delay, accumulating Gate-2 labels, and traversing
both canary stages --- is measured from the onset. Let $t_{\text{flag}}$ be
the time the detector fires and $\hat{t}_d \le t_{\text{flag}}$ its
estimated onset. The retrain then happens at
$t_{\text{rt}} = \max\left(t_{\text{flag}},\; \hat{t}_d + \tau + \lceil
\theta_L W_{\text{train}} \rceil - 1\right)$ (Gate 3 is evaluated at the same
step). The two canary stages follow. Each waits a further $\tau$ for the
labels of the traffic it routes, because it counts \emph{labeled}
instances. Measured from the onset, and ignoring onset-estimation error,
\begin{equation}
L \;\approx\; \max\!\left(L_D,\; \tau + \theta_L W_{\text{train}}\right)
+ \left(\tau + W_1/p_1\right) + \left(\tau + W_2\right),
\label{eq:cycle}
\end{equation}
where $L_D$ is the detection latency. Detection and label accumulation
overlap, so $L_D$ lengthens the cycle only when it exceeds the label wait
(\labelwait{} instances at the frozen defaults). The canary stages add
\canarypart{}. The framework can complete adaptation between drifts only
when $L$ is smaller than the stream's inter-drift interval. The specified
prediction used the first-order form
$\tau + \theta_L W_{\text{train}} + W_1/p_1 + W_2$ (\cyclefirst{}
instances at the frozen defaults), which omits the two canary label waits.
Eq.~(\ref{eq:cycle}) gives \cyclecorr{} whenever $L_D$ is within the label
wait, and Section~\ref{sec:h1results} reports both. Where the condition
fails, RAR cannot finish adapting before the next drift, and we predict
degraded performance. The specified cross-stream prediction is that RAR's
break-even $\rho$ falls as the ratio of inter-drift interval to adaptation
cycle rises.

Second, a \emph{sample penalty}: Gate~2 admits retraining as soon as $\ell$
crosses $\theta_L$, so RAR characteristically trains on
$\approx \theta_L W_{\text{train}}$ labeled instances while ungated
policies train on the full $W_{\text{train}}$. By construction, this is
roughly a $(1-\theta_L)$ sample deficit, traded for the guarantee that
every training instance is post-onset. The trade favors RAR when drift is
real and severe (purity matters) and costs it when drift is absent or the
base learner is sample-hungry. The $W_{\text{train}}$ ablation
(Section~\ref{sec:ablations}) isolates the two effects.

\subsection{Design Trade-offs and Assumptions}
Three assumptions bound the framework's scope, each revisited in
Section~\ref{sec:limitations}. (i) \emph{Labels eventually arrive} with delay
$\tau$; fully unsupervised regimes would require proxy-metric gates.
(ii) \emph{Costs are stationary and known up to the ratio $\rho$}; in
practice $\rho$ must be estimated, and we provide sensitivity analysis rather
than an estimation procedure. (iii) \emph{One challenger at a time}: RAR
maintains a single $M_{\text{new}}$ per cycle. It trades multi-armed
rollout expressiveness for analyzability and for a state machine that maps
onto commodity model registries. We also note an intentional asymmetry: the
gates make \textsc{Retrain} \emph{hard} to fire and \textsc{Alert}
\emph{easy}. This encodes the operational prior that unnecessary retraining
is costlier than a redundant alert. We expect this prior to pay off in the high-$\rho$ regimes, where policy choice matters most.

\section{Experimental Design}
\label{sec:design}

\subsection{Hypotheses (Specified in Advance)}
\label{sec:hypotheses}
We use ``prospectively specified'' rather than ``pre-registered'', because
nothing was filed with an external registry, and we state what can be
verified. Hypotheses H1--H6 below and their verdict branches were written in the
project plan before the base matrix was run. That plan was not under version
control, so the claim that they preceded the matrix rests on the authors' records rather than on a
verifiable timestamp. The two-standard-error verdict rule and the ablation
grid were committed to the version-controlled decisions log before the
earliest committed matrix and ablation results. Two items cannot be shown to
precede their data. The cost-aware baseline's predictions were first
committed in a document that already contained its pilot outcome. The H6
primary metric was first committed together with the first H6 payloads. For
the follow-ups E1--E3 (Section~\ref{sec:followups}), the registration commit
precedes every producing commit. We therefore use three terms. The plan's
hypotheses and rules are \emph{specified}, and their priority rests on the
authors' records. Only the follow-ups E1--E4 are \emph{registered}; their
definitions were committed before execution. Analyses constructed after an
earlier result was known are \emph{post-hoc}. A chronology table built from
the commit history ships with the artifact. Large language models assisted
with writing and debugging the harness, the table-generation and verification
scripts, and parts of the text. The hypotheses, protocol decisions, analyses
and interpretations are the authors' (details under Statements and
Declarations).
\begin{itemize}
\item \textbf{H1 (Dominance).} Full RAR achieves higher net utility than
every binary and single-arm baseline on a majority of stream $\times$ cost
configurations, with rank differences significant under Friedman--Nemenyi
analysis.
\item \textbf{H2 (Rollback value).} On the data-corruption stream, RAR's
rollback arm fires in a non-trivial fraction of trials and prevents a
measurable share of the utility loss incurred by Always-Retrain and Binary
policies.
\item \textbf{H3 (Alert value).} Under label delay ($\tau > 0$), RAR's alert
arm reduces false retrains relative to Binary at equal or better net utility.
\item \textbf{H4 (Drift-type interaction).} The dominant arm varies
systematically with drift type: abrupt drift elevates retrain-arm frequency; gradual drift elevates rollback frequency (premature retraining on incomplete drift); and recurrent drift elevates alert frequency.
\item \textbf{H5 (Canary value).} A two-stage canary reduces the per-deployment cost of bad retrains relative to immediate full deployment, at bounded cost
to good retrains.
\item \textbf{H6 (Post-rollback recovery).} Under a corruption-triggered
rollback followed by genuine abrupt drift, the two-tier hold with emergency
override recovers---retraining once data is clean and drift is severe---where
a single-tier cooldown remains locked, with no loss of oscillation protection under persistent corruption. The primary H6 evaluation metric is the
\emph{mechanism} variable (first committed with the first H6 payloads;
see above): the fraction of trials in which the policy is
still hold-locked when the genuine drift arrives. Net utility is secondary.
Raw retrain lag alone is explicitly not a criterion, because the two
policies reach the corruption with different arm histories and lag can
invert without bearing on the hypothesis.
\end{itemize}

\subsection{Datasets}
\label{sec:datasets}
\begin{sloppypar}
We use eight real public streams, acquired from their providers: the USP Data
Stream Repository for INSECTS and NOAA, River for ELEC2, UCI for Covertype
and Poker Hand, and OpenML for Airlines. Licenses, sources, and checksums are
recorded in the released dataset cards, and no third-party bytes are
redistributed. The INSECTS streams are released by the USP repository
under CC BY 4.0 \citep{souza2020challenges}. NOAA Weather is redistributed
by the same repository from an external source
\citep{ditzler2013incremental}, and the repository's license does not
extend to it. The first four streams are Electricity/ELEC2
\citep{harries1999splice} (45{,}312 instances; real gradual drift) and
INSECTS-Abrupt (imbalanced), INSECTS-Incremental-gradual (imbalanced), and
INSECTS-Incremental-reoccurring (imbalanced) \citep{souza2020challenges}
(355{,}275, 143{,}323, and 452{,}044 instances respectively; documented
change points provide ground-truth onsets on real data). The other four are
Forest Covertype \citep{blackard1999covertype} (581{,}012; slow mixed
drift); NOAA Weather \citep{ditzler2013incremental} (18{,}159; long-horizon
slow drift); Airlines (539{,}383; noisy labels); and Poker Hand
(1{,}025{,}010; minimal drift---a negative control on which rollback should
never fire). We use the imbalanced INSECTS variants as primary because they
preserve the natural temporal order and class prevalence of the original
sensor stream. Balanced variants are curated subsamples that alter stream
dynamics, and a framework motivated by production realism should not be
evaluated on artificially rebalanced data. Streams shorter than the
framework's window structure requires ($\approx$15{,}000 instances given
$W_{\text{train}}$, $\tau$, and canary windows) are excluded by rule.
\end{sloppypar}

Five semi-synthetic streams are constructed by seeded injection into real
streams. Semi-Abrupt (base Electricity) applies a label flip at onset
$u \sim \mathrm{Uniform}[0.4T,
0.8T]$, drawn once with the stream seed and frozen; the onset is 28{,}442
for the ratified stream. Semi-Gradual (base Covertype) applies a linear
label-noise ramp over 5\% of the stream. Semi-Recurrent (base NOAA) has three
drift--recover cycles. Semi-Corrupt (base Airlines) applies feature
corruption from onset onward in two modes. The \emph{blatant} mode resamples
each feature uniformly from its observed domain. The pre-filter must
quarantine this mode, which tests the first defense layer. The \emph{subtle} mode
applies a partial-coverage unit change plus sub-threshold partial nulls. It
is parameterized to evade the pre-filter by construction while still
poisoning a retrain, so it tests the canary-and-rollback layer; this is the
specified H2 instrument. Semi-Rollback-Then-Drift (base Electricity) is the
H6 stream. A transient corruption window triggers an R2 rollback, and genuine
abrupt drift follows after a gap placed inside a naive single-tier cooldown.
Every injected stream carries metadata \{onset, type, width, mode, seed\},
which enables ground-truth recovery-latency and false-retrain metrics.
Injector validation is prospectively specified as a three-part check per
stream: metadata round-trip, a frozen-model accuracy trace breaking at the
recorded onset, and detector sanity. For corruption events, the check adds
quarantine recall and precision against the ground-truth corruption span.
Blatant modes must be caught (recall $\geq 0.5$), subtle modes must evade
(recall $\leq 0.1$), and the clean-region false-positive rate must stay below
the clean-stream quarantine bar. These thresholds are commitments. A subtle
stream the pre-filter catches, or a blatant stream it misses, is an invalid
instrument for H2. Such a stream is re-parameterized before any policy
comparison is run on it. Before the matrix, we performed a column-cardinality
audit of all streams to decide a general identifier-exclusion rule. No
feature met a defensible general criterion; the highest-cardinality candidate
recurs $\approx$80 times per value. All features are therefore retained, and
learner sensitivity to high-cardinality categoricals is reported as a
robustness finding rather than patched per stream. Dataset cards with
licenses, checksums, and the audit accompany the artifact.

\subsection{Policies}
\label{sec:policies}
\begin{sloppypar}
Seven policies share an identical Hoeffding-tree base learner
\citep{domingos2000mining}, in its River implementation
\citep{montiel2021river}, and identical seeds. They are
\textbf{No-Response}; \textbf{Always-Retrain} (every
$W_{\text{train}}$ instances); \textbf{Binary} (ADWIN-triggered retrain,
immediate deployment---the literature-standard policy); \textbf{Alert-Only}
(RAR with retrain and rollback arms disabled); \textbf{Rollback-Only}
(Binary plus two-stage canary, no alert arm); \textbf{Full RAR}; and
\textbf{Cost-Aware Binary}. Cost-Aware Binary is the strongest opponent in
the matrix and the only policy whose decisions read the cost model. It is
Binary plus a single benefit gate that retrains on a detector flag only if
$\hat{\Delta} \cdot \hat{H} \geq \rho$. Here $\hat{\Delta}$ is Full RAR's
own Gate-3 degradation estimate on the same guard-gapped windows, clamped at
zero. This baseline treats it, optimistically, as the per-instance accuracy a
retrain would recover; Gate 3 itself makes no such claim. The horizon
$\hat{H}$ is the observed mean inter-detection interval so far, floored at
$W_{\text{train}}$. $\hat{H}$ is the only free quantity in that gate and is
frozen in advance. A horizon chosen after seeing results would make the
baseline's strength a tunable parameter and every conclusion drawn against
it unfalsifiable. If it proves a poor estimator, that is reported as a
limitation of the baseline, not repaired mid-study. At $\rho \to 0$ the gate
never binds and the policy reduces to Binary. On a fixed detection history
the gate becomes weakly more restrictive as $\rho$ grows. Because retraining
changes the subsequent model and detector histories, however, the realized
number of retrains need not be monotone in $\rho$
(Section~\ref{sec:h2results}). The single-arm policies are included under
the same statistical protocol as H1, but they are not one-factor removals
from Full RAR. Rollback-Only lacks Gates 2--3, the alert arm, and the typed
holds, so Full RAR versus Rollback-Only measures those components jointly.
Alert-Only can never change the deployed model. It serves exactly
No-Response's predictions and differs from it only by alert costs. It is
therefore dominated by construction and serves as a cost control, not as
evidence about alerting. Rollback-Only versus Binary is a clean one-factor
comparison: it adds the two-stage canary to Binary and nothing else. A canary
removal inside Full RAR was run as a registered follow-up
(Section~\ref{sec:followups}). An alert-arm removal was registered but not
run (Section~\ref{sec:followups}).
\end{sloppypar}

\subsection{Metrics and Statistical Protocol}
\label{sec:protocol}
The primary metric is net utility (Eq.~\ref{eq:netutility}). Secondary
metrics are arm firing rates, canary failure rate, and, on semi-synthetic
streams only, recovery latency and false-retrain rate. The last two are
well-defined only where ground-truth onsets exist, a scoping rule we apply
strictly. The configurations are 13 streams $\times$ 6 cost-blind policies
$\times$ 30 seeds $=$ 2{,}340 base-matrix runs, each evaluated analytically
at the four ratified cost regimes ($\rho \in \{0.1,
0.5, 1, 5\}$; Section~\ref{sec:protocol}). To these we add 1{,}560
Cost-Aware Binary runs at explicitly simulated price points, plus a single-tier-cooldown RAR
variant on the Semi-Rollback-Then-Drift stream as the H6 comparison. At each
reported $\rho$, comparison follows \citet{demsar2006statistical}: a
Friedman omnibus test across policies with streams as blocks; Nemenyi
post-hoc tests with critical-difference diagrams; bootstrap 95\% CIs
(1{,}000 resamples) on every reported mean and Cliff's delta for every
pairwise claim, both released with the artifact; the tables in
Section~\ref{sec:results} report paired two-SEM resolution. Cost regimes are not pooled, since they are transformations of
the same runs.

Three decision rules are specified in advance. First, for deterministic base
learners the trial seed varies only the Stage-1 canary routing draw, so
per-cell dispersion measures \emph{routing sensitivity}. We report it as a
finding in its own right, since which marginal challengers happen to commit
can move net utility substantially. Second, a per-cell utility difference
smaller than two standard errors of the mean is reported as
indistinguishable, in either direction. Cells with fewer than three trials
are never assigned a verdict at all. The two-SEM rule assesses uncertainty under routing variation. Because neighboring seeds can share routing draws, the SEM may understate that variation. It is applied per comparison, without correction
for multiplicity, so it is a descriptive criterion rather than a
family-wise test. It is not a minimum-effect criterion either, so a resolved
difference can be operationally negligible. The Poker entries of
Table~\ref{tab:paired} are the extreme case: they are resolved at
$|\Delta U| < 1$ because the seed variance is near zero. Third, every
cost-blind policy's utility is exactly affine in $\rho$
(Section~\ref{sec:formulation}). We therefore compute the $\rho$ dimension
analytically from each run's logged arm counts rather than re-simulating it.
Exactness is asserted by a guard that fails if any policy ever reads the
cost model. The ratified grid $\rho \in \{0.1, 0.5, 1, 5\}$ is reported as
tabulated slices of the full break-even curve. Break-even crossovers are two-directional by construction. Against a costlier-but-more-accurate opponent RAR wins \emph{above} the crossover; against a cheaper-but-less-accurate one it wins \emph{below} it. Here ``costlier'' refers to total action cost $N_r + 0.05 N_a + 0.2 N_b$, not to the cost of one retrain. The direction is
derived from the sign of the accuracy difference. Where both bounds resolve,
we report the resulting \emph{operating window} of $\rho$ in which Full RAR
simultaneously beats Binary, Always-Retrain, and No-Response, with per-seed
ranges. These three are the cost-blind policies whose break-evens are
defined in both directions. Break-evens are never quoted for comparisons
whose utility difference is unresolved. Cost-Aware Binary reads $\rho$, so
its utility is not affine. It is therefore simulated at log-spaced points
$\rho \in \{1, 10, 100, 1000\}$ and excluded from the analytic machinery.
Stage-1 routing is \emph{stratified} as the matrix default
(Section~\ref{sec:rollbackarm}). It routes one instance per block of
$1/p_1$, with a seeded slot per canary episode, matching production
hash-based traffic splitting. We retain i.i.d.\ routing as an ablation cell.
Because the slot seed is the run seed offset by the canary index,
neighboring seeds share slot draws across episodes. Variation across seeds measures routing sensitivity; the seeds are not independent replicates. Alongside mean net utility,
we specify in advance three tail metrics aimed at the catastrophe-avoidance
claims of H2 and H6, outcomes that mean utility detects poorly.
They are worst-seed net utility, maximum rolling accuracy drawdown, and time
spent below the no-skill baseline. All runs are logged with seed and
configuration hash against a version-stamped metrics schema. Before the
matrix consumes any compute, the configuration and metric definitions are frozen
and the freeze is verified by an automated check. The full matrix is
released.

\subsection{Ablations}
\label{sec:ablations}
\begin{sloppypar}
The locked one-factor grid was committed to the decisions log before the
first matrix results. It covers removal of the corruption pre-filter, a guard-gap sweep over $g \in \{0, 100, 200, 400\}$, and a training-window sweep over $W_{\text{train}} \in \{2700, 4000\}$. The training-window sweep isolates the Gate-2 sample penalty from the purity effect. It varies canary margin
$\delta \in \{0.01, 0.03, 0.05\}$, routed fraction
$p_1 \in \{0.1, 0.2, 0.4\}$, and paired windows
$(W_1, W_2)$ $\in$ $\{(150,300)$, $(300,600)$, $(600,1200)\}$, which span the
paired-power range of Appendix~\ref{app:power}. It also varies the gate
thresholds $\theta_L \in \{0.5, 0.7\}$ and $\theta_R \in \{0.03, 0.05, 0.08\}$.
Label delay $\tau \in \{0, 200, 800\}$ crosses $W_1$, so the labeled-window
canary and its timeout path are exercised. The H6-only cooldown sweep
$\kappa_{\text{long}} \in \{2500, 5000, 10000\}$ tests whether the two-tier
result is an artifact of one cooldown value. The grid also includes i.i.d.\
versus stratified routing and emergency threshold
$\Delta_{\text{crit}} \in \{2, 3\} \times \theta_R$. Three audit cells test configuration choices the authors had approved without measurement:
$z_{\text{commit}} \in \{0, 1\}$, the categorical-exemption cardinality
bound $\in \{2, 20\}$, and detector warmup $\in \{30, 500\}$. The
collapse-monitor thresholds are accepted conditionally. If disagreement
aborts occur in at least 5\% of matrix runs, a sensitivity cell over the collapse, disagreement, and incumbent-variation thresholds is triggered. The condition is specified here so the choice
cannot be made after seeing results.
\end{sloppypar}

\subsection{Registered Follow-ups}
\label{sec:followups}
During our review of an earlier draft, we identified four gaps in the
experimental evaluation. H2 had not been
run on its specified instrument, H5 had no canary removal inside Full RAR,
all policy results used one learner, and the alert arm was not isolated. We
registered four follow-ups (E1--E4) in the decisions log before running
them. The stream, policies, seeds (0--29), routing, $\tau$, verdict rules,
and reporting obligations were frozen in advance. The three that ran (E1, E2a, E3) used the current API, whose Full RAR reproduces the frozen base-matrix Full RAR cell for cell
(\apiequiv{} of \apiequivtotal{} identical net utilities), so the follow-ups
are comparable with the base matrix. \textbf{E1} completes the omitted H2
experiment on the subtle stream with the six cost-blind policies (180 runs).
\textbf{E2a} evaluates Full RAR with the canary
switched off and nothing else, on all 13 streams (390 runs, paired with 390
re-runs of Full RAR). \textbf{E3} repeats four policies with a Gaussian
naive-Bayes learner on five streams (600 runs), as a robustness panel with no
hypothesis. \textbf{E2b}, a matched per-deployment test of H5, and
\textbf{E4}, an alert-arm removal, were stopped at registered pre-launch
checks and are reported as such (Section~\ref{sec:h5results}).

\section{Results}
\label{sec:results}
Results in this section come from the frozen base matrix unless they are
marked as a registered follow-up (E1--E3, Section~\ref{sec:followups}), the
cost-aware matrix, or the H6 follow-up. The base matrix has 13
streams $\times$ 6 cost-blind policies $\times$ 30 seeds $=$ 2{,}340 cells,
and every cell has audit status PASS. All base-matrix cells share one API version (1.1) and one configuration hash (\texttt{4c169401ccb9}). The matrix uses a Hoeffding-tree
base learner, $\tau = 800$, and stratified Stage-1 routing. For every cell,
the ledger identity
$n_{\text{correct}} = \mathrm{round}(\text{accuracy} \times n_{\text{instances}})$
and $U(1) = n_{\text{correct}} - (N_r + 0.05 N_a + 0.2 N_b)$ was re-verified
before any table was built. Every table below is generated from the cell
records by the released analysis script. Utility at other $\rho$ is the exact
affine recomputation of Section~\ref{sec:design}. \emph{Verdict rule:} a paired per-seed difference is \emph{resolved} ($^{*}$) when the absolute value of its mean exceeds two standard errors of the paired differences over 30 seeds. Unmarked differences are reported as
indistinguishable in either direction. Break-even crossovers are reported
only when all 30 seeds agree on their direction. The reported value is the mean of the 30 per-seed crossovers, which can differ slightly from the crossover of the mean utilities, and the seed range is in brackets. Otherwise the comparison is \emph{disputed}, and the seed
composition is printed. Every specified quantity is reported below. Where a
specified quantity is not estimable from the frozen ledger, the section says
so rather than substituting a proxy.

\subsection{H1 --- Overall Policy Comparison}
\label{sec:h1results}
\begin{table}[t]
\caption{Mean net utility at the ratified $\rho = 1$ over 30 seeds; best per
stream in bold; Full RAR's seed standard deviation in small type.
Rollback-Only also varies by seed through its canary routing (its dispersion
is reported in Table~\ref{tab:tail}); the four canary-free policies are
deterministic given the stream.}
\label{tab:mainresults}
\centering\footnotesize\setlength{\tabcolsep}{3pt}
\begin{tabular}{@{}lrrrrrr@{}}
\toprule
Stream & No-Resp. & Always-Retr. & Binary & Alert-Only & Rollback-Only & Full RAR \\
\midrule
ELEC2 & 32,430 & \textbf{33,157} & 32,538 & 32,428 & 32,292 & 32,623 {\scriptsize$\pm$1,233} \\
Covertype & 237,095 & 391,981 & \textbf{401,208} & 237,072 & 382,613 & 372,416 {\scriptsize$\pm$4,777} \\
Poker & \textbf{513,697} & 513,318 & \textbf{513,697} & \textbf{513,697} & \textbf{513,697} & \textbf{513,697 {\scriptsize$\pm$0}} \\
Airlines & 289,503 & 324,869 & \textbf{330,968} & 289,483 & 321,438 & 302,233 {\scriptsize$\pm$4,305} \\
NOAA & 12,454 & \textbf{12,585} & 12,446 & 12,453 & 12,505 & 12,452 {\scriptsize$\pm$0} \\
INS-abrupt & 171,634 & \textbf{194,122} & 187,024 & 171,621 & 174,641 & 189,793 {\scriptsize$\pm$4,902} \\
INS-gradual & 57,671 & \textbf{82,919} & 73,128 & 57,665 & 64,657 & 64,310 {\scriptsize$\pm$41} \\
INS-recurr. & 78,499 & \textbf{245,533} & 219,997 & 78,477 & 190,362 & 200,903 {\scriptsize$\pm$14,687} \\
S-Abrupt & 24,714 & \textbf{32,335} & 31,093 & 24,712 & 28,999 & 29,868 {\scriptsize$\pm$614} \\
S-Gradual & 209,490 & 386,761 & \textbf{392,852} & 209,466 & 377,088 & 364,832 {\scriptsize$\pm$4,764} \\
S-Recurrent & 11,002 & 11,180 & \textbf{11,299} & 11,001 & 11,045 & 10,985 {\scriptsize$\pm$13} \\
S-Corrupt & 291,319 & 313,040 & \textbf{314,664} & 291,298 & 306,030 & 291,868 {\scriptsize$\pm$2,293} \\
S-RB-Drift & 24,956 & \textbf{31,687} & 31,093 & 24,954 & 28,087 & 31,300 {\scriptsize$\pm$16} \\
\bottomrule
\end{tabular}

\end{table}

\begin{table}[t]
\caption{Full RAR minus baseline, paired by seed, at the ratified $\rho = 1$
and $\rho = 5$. $^{*}$: resolved at two standard errors, which indicates
certainty under routing variation, not a meaningful effect (Poker's $-$0.1$^{*}$ and $-$0.5$^{*}$ are
sub-unit differences with zero variance).}
\label{tab:paired}
\centering\footnotesize\setlength{\tabcolsep}{2.5pt}
\begin{tabular}{@{}lrrrrrrrr@{}}
\toprule
Stream & \multicolumn{2}{c}{vs.\ Always-Retr.} & \multicolumn{2}{c}{vs.\ Binary} & \multicolumn{2}{c}{vs.\ Rollback-Only} & \multicolumn{2}{c}{vs.\ No-Resp.} \\
 & $\Delta U(1)$ & $\Delta U(5)$ & $\Delta U(1)$ & $\Delta U(5)$ & $\Delta U(1)$ & $\Delta U(5)$ & $\Delta U(1)$ & $\Delta U(5)$ \\
\midrule
ELEC2 & $-$534$^{*}$ & $-$492$^{*}$ & +85 & +171 & +331 & +349 & +193 & +171 \\
Covertype & $-$19,565$^{*}$ & $-$19,023$^{*}$ & $-$28,792$^{*}$ & $-$27,990$^{*}$ & $-$10,197$^{*}$ & $-$10,033$^{*}$ & +135,321$^{*}$ & +135,003$^{*}$ \\
Poker & +379$^{*}$ & +1,894$^{*}$ & $-$0.1$^{*}$ & $-$0.5$^{*}$ & $-$0.1$^{*}$ & $-$0.5$^{*}$ & $-$0.1$^{*}$ & $-$0.5$^{*}$ \\
Airlines & $-$22,636$^{*}$ & $-$21,946$^{*}$ & $-$28,735$^{*}$ & $-$28,673$^{*}$ & $-$19,205$^{*}$ & $-$19,186$^{*}$ & +12,730$^{*}$ & +12,624$^{*}$ \\
NOAA & $-$133$^{*}$ & $-$118$^{*}$ & +6$^{*}$ & +24$^{*}$ & $-$53$^{*}$ & $-$49$^{*}$ & $-$2$^{*}$ & $-$12$^{*}$ \\
INS-abrupt & $-$4,329$^{*}$ & $-$3,845$^{*}$ & +2,769$^{*}$ & +2,777$^{*}$ & +15,152$^{*}$ & +15,146$^{*}$ & +18,159$^{*}$ & +18,119$^{*}$ \\
INS-gradual & $-$18,609$^{*}$ & $-$18,431$^{*}$ & $-$8,818$^{*}$ & $-$8,832$^{*}$ & $-$346$^{*}$ & $-$364$^{*}$ & +6,639$^{*}$ & +6,609$^{*}$ \\
INS-recurr. & $-$44,630$^{*}$ & $-$44,063$^{*}$ & $-$19,094$^{*}$ & $-$19,111$^{*}$ & +10,541 & +10,523 & +122,404$^{*}$ & +122,303$^{*}$ \\
S-Abrupt & $-$2,467$^{*}$ & $-$2,425$^{*}$ & $-$1,225$^{*}$ & $-$1,147$^{*}$ & +868$^{*}$ & +885$^{*}$ & +5,154$^{*}$ & +5,132$^{*}$ \\
S-Gradual & $-$21,929$^{*}$ & $-$21,383$^{*}$ & $-$28,020$^{*}$ & $-$27,186$^{*}$ & $-$12,256$^{*}$ & $-$12,093$^{*}$ & +155,342$^{*}$ & +155,028$^{*}$ \\
S-Recurrent & $-$195$^{*}$ & $-$185$^{*}$ & $-$314$^{*}$ & $-$308$^{*}$ & $-$60$^{*}$ & $-$60$^{*}$ & $-$17$^{*}$ & $-$31$^{*}$ \\
S-Corrupt & $-$21,172$^{*}$ & $-$20,486$^{*}$ & $-$22,796$^{*}$ & $-$22,738$^{*}$ & $-$14,162$^{*}$ & $-$14,181$^{*}$ & +549 & +439 \\
S-RB-Drift & $-$387$^{*}$ & $-$348$^{*}$ & +207$^{*}$ & +274$^{*}$ & +3,214$^{*}$ & +3,220$^{*}$ & +6,344$^{*}$ & +6,319$^{*}$ \\
\bottomrule
\end{tabular}

\end{table}

\textbf{H1's result falls in its prospectively specified partial-support branch.} At the ratified
$\rho = 1$ (Table~\ref{tab:mainresults}), Always-Retrain or Binary has the
highest mean utility on every stream. There is no stream on which Full RAR
beats all four non-trivial baselines simultaneously with resolved differences
(\numwinsrhoresolvedall{} of 13). At $\rho = 1$, Full RAR beats Binary
(resolved) on three streams: NOAA, INSECTS-abrupt, and the
rollback-then-drift stream (Table~\ref{tab:paired}). It beats Rollback-Only
on three streams (INSECTS-abrupt, semi-abrupt, rollback-then-drift) and loses
to it on eight; all eleven differences are resolved, and the remaining two
(ELEC2, INSECTS-recurrent) are not. One of the eight losses is Poker's
$-0.1$, which is resolved only because its variance is zero and is
negligible in size.
Across all \numpwtotal{} Full-RAR pairwise effects (13 streams $\times$ 5
comparators $\times$ 5 price points, $\rho \in \{1, 5, 10, 100, 200\}$), \numpwres{} are resolved under the
specified rule: \numpwfavrar{} favor Full RAR and \numpwfavcomp{} favor a
comparator. Bootstrap 95\% intervals and Cliff's $\delta$ for every pair are
released with the artifact. Against Always-Retrain, Full RAR loses at
$\rho \le 5$ on every stream except the Poker negative control, where
Always-Retrain pays for 379 useless retrains. That is the prediction of
Section~\ref{sec:formulation}. At those prices a retrain costs one correct
prediction in tens of thousands, and acting constantly is near-optimal by
construction. The picture inverts as retraining becomes expensive. By mean
utility at $\rho = 100$, Full RAR is ahead of Binary on five streams (ELEC2,
NOAA, INSECTS-abrupt, semi-abrupt, rollback-then-drift) and ahead of
Always-Retrain on six.

\begin{table}[t]
\caption{Friedman mean ranks over the 13 stream blocks (rank 1 = best) at each
$\rho$ slice, with the tie-corrected omnibus test. Nemenyi critical difference at
$\alpha = 0.05$, $k = 6$, $N = 13$: $\mathrm{CD} = \cdvalue$.}
\label{tab:cd}
\centering\footnotesize\setlength{\tabcolsep}{3pt}
\begin{tabular}{@{}lrrrrrrr@{}}
\toprule
$\rho$ & No-Resp. & Always-Retr. & Binary & Alert-Only & Rollback-Only & Full RAR & Friedman \\
\midrule
0.1 & 4.46 & 1.77 & 2.15 & 5.58 & 3.38 & 3.65 & $p<0.001$ \\
0.5 & 4.46 & 1.77 & 2.15 & 5.58 & 3.38 & 3.65 & $p<0.001$ \\
1 & 4.46 & 1.77 & 2.15 & 5.58 & 3.38 & 3.65 & $p<0.001$ \\
5 & 4.42 & 1.77 & 2.19 & 5.58 & 3.38 & 3.65 & $p<0.001$ \\
100 & 3.77 & 2.92 & 2.69 & 4.88 & 3.31 & 3.42 & $p=0.045$ \\
200 & 3.54 & 4.00 & 3.15 & 4.65 & 2.69 & 2.96 & $p=0.077$ \\
\bottomrule
\end{tabular}

\end{table}

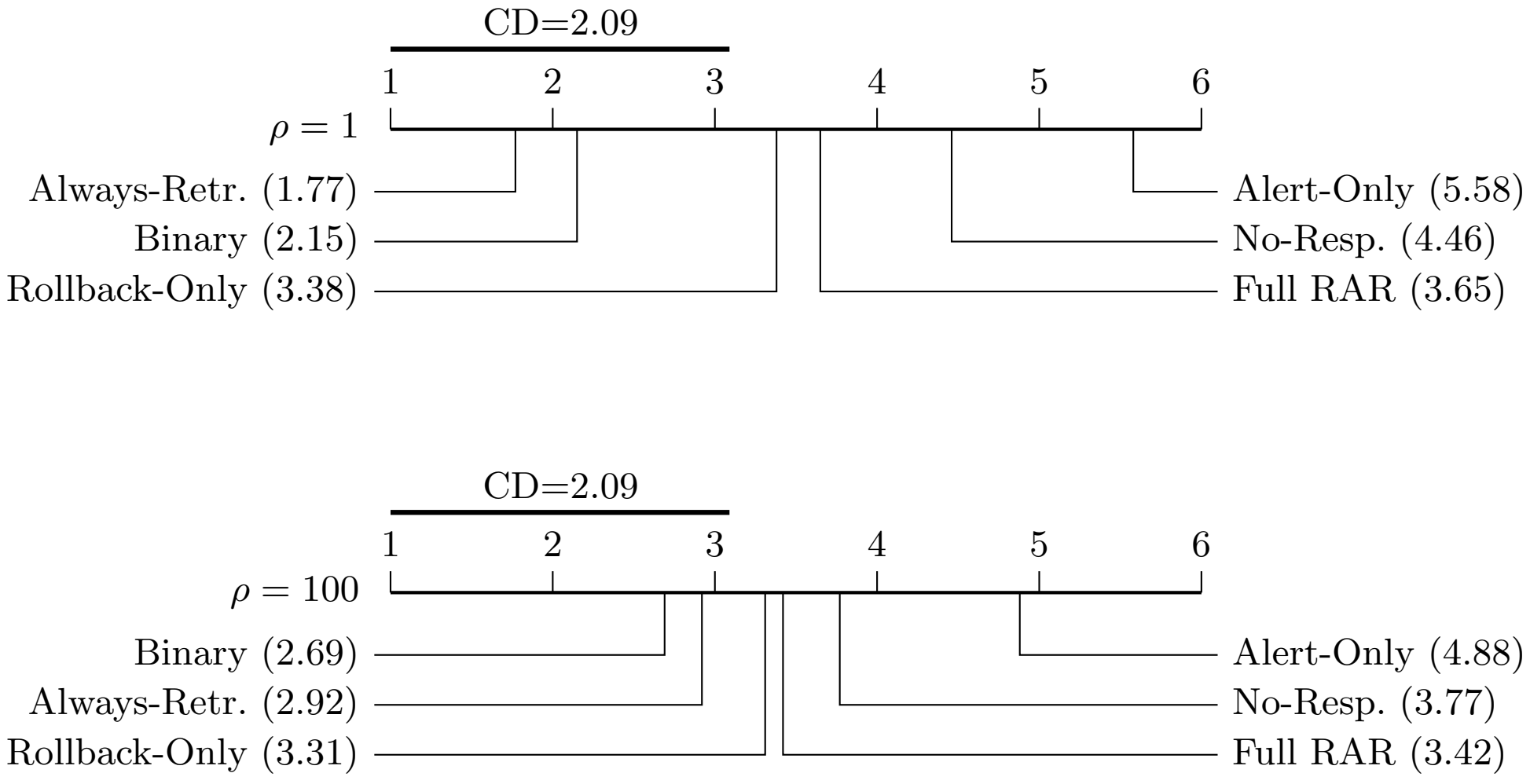
\begin{figure}[t]
\centering
{\tikzset{every picture/.style={xscale=1.3,yscale=1.5}}
\begin{tikzpicture}[x=1.05cm,y=1cm]
\draw[thick] (1,0) -- (6,0);
\draw (1,0) -- (1,0.12); \node[above,font=\scriptsize] at (1,0.12) {1};
\draw (2,0) -- (2,0.12); \node[above,font=\scriptsize] at (2,0.12) {2};
\draw (3,0) -- (3,0.12); \node[above,font=\scriptsize] at (3,0.12) {3};
\draw (4,0) -- (4,0.12); \node[above,font=\scriptsize] at (4,0.12) {4};
\draw (5,0) -- (5,0.12); \node[above,font=\scriptsize] at (5,0.12) {5};
\draw (6,0) -- (6,0.12); \node[above,font=\scriptsize] at (6,0.12) {6};
\node[left,font=\scriptsize] at (0.9,0) {$\rho=1$};
\draw (1.77,0) -- (1.77,-0.35) -- (0.9,-0.35); \node[left,font=\scriptsize] at (0.9,-0.35) {Always-Retr. (1.77)};
\draw (2.15,0) -- (2.15,-0.63) -- (0.9,-0.63); \node[left,font=\scriptsize] at (0.9,-0.63) {Binary (2.15)};
\draw (3.38,0) -- (3.38,-0.91) -- (0.9,-0.91); \node[left,font=\scriptsize] at (0.9,-0.91) {Rollback-Only (3.38)};
\draw (3.65,0) -- (3.65,-0.91) -- (6.1,-0.91); \node[right,font=\scriptsize] at (6.1,-0.91) {Full RAR (3.65)};
\draw (4.46,0) -- (4.46,-0.63) -- (6.1,-0.63); \node[right,font=\scriptsize] at (6.1,-0.63) {No-Resp. (4.46)};
\draw (5.58,0) -- (5.58,-0.35) -- (6.1,-0.35); \node[right,font=\scriptsize] at (6.1,-0.35) {Alert-Only (5.58)};
\draw[very thick] (1,0.45) -- (3.09,0.45); \node[above,font=\scriptsize] at (2.05,0.45) {CD=2.09};
\draw[thick] (1,-2.6) -- (6,-2.6);
\draw (1,-2.6) -- (1,-2.48); \node[above,font=\scriptsize] at (1,-2.48) {1};
\draw (2,-2.6) -- (2,-2.48); \node[above,font=\scriptsize] at (2,-2.48) {2};
\draw (3,-2.6) -- (3,-2.48); \node[above,font=\scriptsize] at (3,-2.48) {3};
\draw (4,-2.6) -- (4,-2.48); \node[above,font=\scriptsize] at (4,-2.48) {4};
\draw (5,-2.6) -- (5,-2.48); \node[above,font=\scriptsize] at (5,-2.48) {5};
\draw (6,-2.6) -- (6,-2.48); \node[above,font=\scriptsize] at (6,-2.48) {6};
\node[left,font=\scriptsize] at (0.9,-2.6) {$\rho=100$};
\draw (2.69,-2.6) -- (2.69,-2.95) -- (0.9,-2.95); \node[left,font=\scriptsize] at (0.9,-2.95) {Binary (2.69)};
\draw (2.92,-2.6) -- (2.92,-3.2300000000000004) -- (0.9,-3.2300000000000004); \node[left,font=\scriptsize] at (0.9,-3.2300000000000004) {Always-Retr. (2.92)};
\draw (3.31,-2.6) -- (3.31,-3.5100000000000002) -- (0.9,-3.5100000000000002); \node[left,font=\scriptsize] at (0.9,-3.5100000000000002) {Rollback-Only (3.31)};
\draw (3.42,-2.6) -- (3.42,-3.5100000000000002) -- (6.1,-3.5100000000000002); \node[right,font=\scriptsize] at (6.1,-3.5100000000000002) {Full RAR (3.42)};
\draw (3.77,-2.6) -- (3.77,-3.2300000000000004) -- (6.1,-3.2300000000000004); \node[right,font=\scriptsize] at (6.1,-3.2300000000000004) {No-Resp. (3.77)};
\draw (4.88,-2.6) -- (4.88,-2.95) -- (6.1,-2.95); \node[right,font=\scriptsize] at (6.1,-2.95) {Alert-Only (4.88)};
\draw[very thick] (1,-2.15) -- (3.09,-2.15); \node[above,font=\scriptsize] at (2.05,-2.15) {CD=2.09};
\end{tikzpicture}
}
\caption{Critical-difference diagrams at $\rho = 1$ (top) and $\rho = 100$
(bottom); policies within one CD of each other are not separable. Full RAR is within one CD of every comparator at both prices.}
\label{fig:cd}
\end{figure}

The rank analysis says the same thing more precisely (Table~\ref{tab:cd},
Figure~\ref{fig:cd}). At $\rho \le 5$ the Friedman omnibus rejects a common
ranking ($p < 0.001$). Alert-Only and No-Response at the bottom and
Always-Retrain and Binary at the top drive the rejection. Full RAR sits
mid-table at mean rank 3.65, within one critical difference (2.09) of Binary,
Rollback-Only, and No-Response alike. At $\rho = 100$ the ranks compress into
a 2.2-rank band, and the omnibus barely rejects ($p = \friedmanphund$,
tie-corrected). At $\rho = 200$ it no longer does ($p = \friedmanptwohund$
over all 13 streams; Table~\ref{tab:cd}). The loss of omnibus significance at high cost is consistent with the paper's main result. As retraining becomes expensive,
there is no consistent cross-stream ordering of policies at all, and rank
becomes a property of the stream. This is why the deliverable of this paper
is an operating map rather than a winner.

The thirteen blocks are not independent. Twelve of them come from five base
datasets, and the 30 seeds vary only canary routing, so they measure routing
sensitivity rather than independent draws of the stream. We therefore repeat
the rank analysis with one block per base dataset, and again leaving out each
base dataset in turn (Table~\ref{tab:lobo}). The conclusions hold. At
$\rho = 1$ the omnibus still rejects with six base-dataset blocks
($p = \fampone$) and in every leave-one-out fold (largest $p$ \lobopmaxone).
At $\rho = 100$ it no longer rejects with six blocks ($p = \famphund$) and in
\lobodissolvehund{} of six folds. Full RAR is separated from no comparator in
\lobonotsep{} of the \lobototal{} grouped and leave-one-out analyses. The
exception is that it separates \emph{above} Alert-Only at $\rho = 1$ when
NOAA is left out.

\afterpage{\begin{landscape}
\begin{table}
\caption{Rank analysis grouped by base dataset: mean ranks with all 13
stream blocks, with one block per base dataset (within-family ranks averaged,
then re-ranked), and leaving each base dataset out. CD is the Nemenyi
critical difference for that $N$.}
\label{tab:lobo}
\centering\scriptsize\setlength{\tabcolsep}{2.5pt}
\begin{tabular}{@{}rlrrrrrrrrr@{}}
\toprule
$\rho$ & Blocks & $N$ & No-Resp. & Always-Retr. & Binary & Alert-Only & Rollback-Only & Full RAR & $p$ & CD \\
\midrule
1 & all 13 streams & 13 & 4.46 & 1.77 & 2.15 & 5.58 & 3.38 & 3.65 & $<$0.001 & 2.09 \\
1 & 6 base-dataset blocks & 6 & 4.17 & 2.17 & 2.08 & 5.58 & 3.08 & 3.92 & 0.007 & 3.08 \\
1 & without Electricity & 10 & 4.40 & 2.00 & 2.00 & 5.55 & 3.00 & 4.05 & $<$0.001 & 2.38 \\
1 & without Covertype & 11 & 4.36 & 1.73 & 2.36 & 5.50 & 3.45 & 3.59 & $<$0.001 & 2.27 \\
1 & without Airlines & 11 & 4.36 & 1.73 & 2.36 & 5.50 & 3.45 & 3.59 & $<$0.001 & 2.27 \\
1 & without NOAA & 11 & 4.64 & 1.82 & 1.91 & 5.77 & 3.55 & 3.32 & $<$0.001 & 2.27 \\
1 & without INSECTS & 10 & 4.30 & 2.00 & 2.10 & 5.45 & 3.30 & 3.85 & $<$0.001 & 2.38 \\
1 & without Poker & 12 & 4.67 & 1.42 & 2.17 & 5.67 & 3.50 & 3.58 & $<$0.001 & 2.18 \\
100 & all 13 streams & 13 & 3.77 & 2.92 & 2.69 & 4.88 & 3.31 & 3.42 & 0.045 & 2.09 \\
100 & 6 base-dataset blocks & 6 & 3.33 & 3.17 & 2.67 & 5.08 & 3.25 & 3.50 & 0.306 & 3.08 \\
100 & without Electricity & 10 & 3.80 & 3.10 & 2.30 & 4.95 & 3.00 & 3.85 & 0.034 & 2.38 \\
100 & without Covertype & 11 & 3.55 & 3.18 & 2.82 & 4.68 & 3.45 & 3.32 & 0.272 & 2.27 \\
100 & without Airlines & 11 & 3.64 & 2.91 & 3.00 & 4.77 & 3.55 & 3.14 & 0.183 & 2.27 \\
100 & without NOAA & 11 & 4.27 & 2.45 & 2.36 & 5.41 & 3.18 & 3.32 & $<$0.001 & 2.27 \\
100 & without INSECTS & 10 & 3.40 & 3.30 & 2.90 & 4.55 & 3.20 & 3.65 & 0.452 & 2.38 \\
100 & without Poker & 12 & 3.92 & 2.67 & 2.75 & 4.92 & 3.42 & 3.33 & 0.036 & 2.18 \\
\bottomrule
\end{tabular}

\end{table}
\begin{table}
\caption{Operating windows. Crossovers against costlier-but-more-accurate baselines are lower bounds ($\rho >$), against cheaper-but-less-accurate ones upper bounds ($\rho <$); each bound is the mean of the per-seed crossovers; ``always''/``never'' denote dominance in one
direction on every seed; brackets give the seed range; for disputed cells
the small type counts seeds by outcome (ab: above the crossover, al: always,
be: below, ne: never). The window is the
interval in which Full RAR beats all three simultaneously; it is marked
disputed if any bound is.}
\label{tab:window}
\centering\scriptsize\setlength{\tabcolsep}{2pt}
\begin{tabular}{@{}lllll@{}}
\toprule
Stream & vs.\ Binary & vs.\ Always-Retr. & vs.\ No-Resp. & Window \\
\midrule
ELEC2 & disputed {\scriptsize 10\,ab 20\,al} & disputed {\scriptsize 10\,ab 20\,al} & disputed {\scriptsize 20\,be 10\,ne} & disputed \\
Covertype & $\rho>144.6$ {\scriptsize [104--198]} & $\rho>145.5$ {\scriptsize [86--225]} & $\rho<1,706.4$ {\scriptsize [1,482--1,817]} & $(146,\,1,706)$ \\
Poker & never & always & never & none \\
Airlines & $\rho>2,011.3$ {\scriptsize [704--3,557]} & $\rho>132.2$ {\scriptsize [53--162]} & $\rho<493.9$ {\scriptsize [284--1,298]} & none \\
NOAA & always & $\rho>37.0$ {\scriptsize [37--37]} & never & none \\
INS-abrupt & disputed {\scriptsize 8\,ab 16\,al 6\,be} & disputed {\scriptsize 20\,ab 10\,al} & $\rho<1,806.4$ {\scriptsize [1,318--2,428]} & disputed \\
INS-gradual & never & $\rho>419.2$ {\scriptsize [418--421]} & $\rho<886.3$ {\scriptsize [874--892]} & none \\
INS-recurr. & disputed {\scriptsize 7\,be 23\,ne} & $\rho>315.7$ {\scriptsize [131--400]} & $\rho<4,865.8$ {\scriptsize [4,158--6,169]} & disputed \\
S-Abrupt & $\rho>64.2$ {\scriptsize [1--90]} & $\rho>237.1$ {\scriptsize [113--296]} & $\rho<960.7$ {\scriptsize [772--1,303]} & $(237,\,961)$ \\
S-Gradual & $\rho>135.4$ {\scriptsize [99--196]} & $\rho>161.7$ {\scriptsize [106--255]} & $\rho<1,984.1$ {\scriptsize [1,797--2,130]} & $(162,\,1,984)$ \\
S-Recurrent & $\rho>211.4$ {\scriptsize [133--234]} & $\rho>78.5$ {\scriptsize [61--88]} & disputed {\scriptsize 5\,be 25\,ne} & disputed \\
S-Corrupt & $\rho>1,589.3$ {\scriptsize [848--2,027]} & $\rho>124.5$ {\scriptsize [71--142]} & disputed {\scriptsize 20\,be 10\,ne} & disputed \\
S-RB-Drift & always & $\rho>40.7$ {\scriptsize [38--44]} & $\rho<1,013.2$ {\scriptsize [1,003--1,020]} & $(41,\,1,013)$ \\
\bottomrule
\end{tabular}

\end{table}
\end{landscape}}

\textbf{Operating windows} (Table~\ref{tab:window}). Four streams yield a fully
resolved two-sided window in which Full RAR beats Binary, Always-Retrain, and
No-Response at once: rollback-then-drift ($41 < \rho < 1{,}013$),
Covertype ($146$--$1{,}706$), semi-gradual ($162$--$1{,}984$), and semi-abrupt
($237$--$961$). Each window is bounded below by the price at which gating
pays for its overhead and above by the price at which doing nothing wins. On
Airlines, INSECTS-gradual, NOAA and Poker no window exists. On five streams a
bound is disputed across seeds, and no window is claimed. The lower bounds
sit at $\rho \approx 40$--$240$, one to two orders of magnitude above the
ratified grid; Section~\ref{sec:limitations} discusses what that says about
the grid.

\afterpage{\begin{landscape}
\begin{table}
\caption{Latency feasibility (Section~\ref{sec:framework}): each stream's
available inter-drift interval divided by the specified first-order cycle
(\cyclefirst{} instances, Ratio) and by the cycle including both canary label
waits (\cyclecorr{}, Ratio$'$), the specified assessment, and the sign of Full RAR's paired $\Delta U(1)$ against
Binary ($^{*}$ resolved). The prediction is tested only on rows with a
determinable interval.}
\label{tab:feas}
\centering\scriptsize\setlength{\tabcolsep}{2.5pt}
\begin{tabular}{@{}lp{0.30\columnwidth}rrll@{}}
\toprule
Stream & Change structure & Ratio & Ratio$'$ & Assessment & RAR vs.\ Binary \\
\midrule
ELEC2 & real, empirical detector interval $\sim$1,678 & 0.35 & 0.26 & INFEASIBLE & + \\
Covertype & real stream; documented spacing unavailable & -- & -- & UNDETERMINED & $-$$^{*}$ \\
Poker & no documented drift; negative control & -- & -- & NOT APPLICABLE & $-$$^{*}$ \\
Airlines & real stream; documented spacing unavailable & -- & -- & UNDETERMINED & $-$$^{*}$ \\
NOAA & clean 18,159-row stream; no documented drift & 3.79 total cycles & 2.84 total cycles & NOT A DRIFT TEST & +$^{*}$ \\
INS-abrupt & documented points; minimum gap 25,497 & 5.32 & 3.99 & FEASIBLE & +$^{*}$ \\
INS-gradual & documented gradual transition; no stationary gap & -- & -- & INFEASIBLE & $-$$^{*}$ \\
INS-recurr. & incremental/reoccurring change; no stationary gap & -- & -- & INFEASIBLE & $-$$^{*}$ \\
S-Abrupt & permanent label drift; 16,870 rows after onset & 3.52 & 2.64 & FEASIBLE & $-$$^{*}$ \\
S-Gradual & 29,051-row transition then long post-transition tail & mixed & -- & MIXED & $-$$^{*}$ \\
S-Recurrent & 2,253 rows between reversals & 0.47 & 0.35 & INFEASIBLE & $-$$^{*}$ \\
S-Corrupt & permanent feature corruption, not recurrent target drift & -- & -- & NOT APPLICABLE & $-$$^{*}$ \\
S-RB-Drift & true label drift; 14,104 rows after onset & 2.95 & 2.21 & FEASIBLE & +$^{*}$ \\
\bottomrule
\end{tabular}

\end{table}
\end{landscape}}

\textbf{The results agree with the specified latency prediction in both directions}
(Table~\ref{tab:feas}). Of the three FEASIBLE streams, Full RAR beats Binary
(resolved) at $\rho = 1$ on two (INSECTS-abrupt, rollback-then-drift) and
crosses it at $\rho > 64$ on the third. Of the four INFEASIBLE streams, it is
behind Binary on three (resolved) and indistinguishable from it on the fourth
(ELEC2). UNDETERMINED, MIXED, NOT-APPLICABLE and NOT-A-DRIFT-TEST rows are
excluded by the rule stated in advance. Recomputing the ratios with the
corrected cycle of Eq.~(\ref{eq:cycle}) lowers every ratio by a quarter and
changes no assessment. By Eq.~(\ref{eq:cycle}), the three FEASIBLE verdicts
survive any detection latency below \feaslatency{} instances. INSECTS-abrupt
is Full RAR's strongest real-stream result in the matrix. It is a
355K-instance real stream with documented change points a minimum of
25{,}497 instances apart, ratio 5.32. INSECTS-gradual and -recurrent, where
change is continuous or rapidly cycling, are among Full RAR's weakest. This
pattern is consistent with the framework's feasibility condition. Even so, it
is evidence of association rather than a tested prediction: there are three
FEASIBLE and four INFEASIBLE evaluable streams, and one FEASIBLE result
reverses under a second learner (Section~\ref{sec:e3}).
Section~\ref{sec:erroranalysis} shows that the Gate-3 estimate is
correspondingly clean on abrupt and blind on recurrent streams.

\subsection{H2 --- Rollback on Corruption}
\label{sec:h2results}
The pre-filter behaves as specified. On the blatant corruption stream, $q_t$
recall is 0.999 at precision 1.000 with zero clean-region false positives. On
the transient corruption block of the rollback-then-drift stream, recall is
0.958 at precision 0.935 with a 0.37\% clean false-positive rate (below the
0.5\% bar). On the subtle stream recall is 0.005, i.e.\ it evades the
pre-filter by construction while remaining detectable downstream (ADWIN lag
$+4{,}167$) and toxic ($+4.1$ points of error). On rollback-then-drift, R2
fires once per run on every seed, and the quarantine rate is 5.12\% against a
5.00\% injected block. Full RAR beats Rollback-Only and Binary by
$+3{,}214$ and $+207$ at $\rho = 1$, both resolved (Table~\ref{tab:paired}),
though it differs from both in more than the corruption arm
(Section~\ref{sec:policies}). Its accuracy seed-dispersion is \numrbdsd{}
against Rollback-Only's \numrbdrosd{}.

\textbf{H2 is not supported on its specified instrument.} H2 was specified
on the subtle corruption stream, which evades the pre-filter and so exercises
the canary-and-rollback layer. The base-matrix launch package classified that
stream as a diagnostic and ran the blatant stream in its place. We record
this as a deviation and report the registered experiment separately as E1
(Table~\ref{tab:e1}). E1 was completed after the rest of the results were
known and run under the frozen definitions. The pre-filter catches
\eonerecall{} of the subtle corruption, as designed. The rollback arm fires:
Full RAR has an R1 or R3 rollback after the corruption onset on \eonefire{}
of 30 seeds, above E1's registered threshold of 10. But it does not prevent
the loss the hypothesis names. At $\rho = 1$, Full RAR minus Binary is
\eonebin{} and minus Always-Retrain is \eonear{}, negative on all 30 seeds
each; minus Rollback-Only it is \eonero{}. It is ahead only of the two
policies that never retrain. Under E1's registered verdict rule, H2 is not
supported.

\begin{table}[t]
\caption{E1: H2 on its specified subtle-corruption stream, 30 paired seeds,
$\rho = 1$ ($^{*}$ resolved). Rollbacks per run include all causes.}
\label{tab:e1}
\centering\scriptsize\setlength{\tabcolsep}{2.5pt}
\begin{tabular}{@{}lrrrrr@{}}
\toprule
Policy & Mean $U(1)$ & Full RAR $-$ policy & 2$\cdot$SEM & $+$/$-$ seeds & Rollbacks/run \\
\midrule
No-Resp. & 291,741 & +9,777$^{*}$ & 1,416 & 30/0 & 0.0 \\
Always-Retr. & 324,879 & $-$23,361$^{*}$ & 1,416 & 0/30 & 0.0 \\
Binary & 327,337 & $-$25,819$^{*}$ & 1,416 & 0/30 & 0.0 \\
Alert-Only & 291,721 & +9,797$^{*}$ & 1,416 & 30/0 & 0.0 \\
Rollback-Only & 319,573 & $-$18,055$^{*}$ & 1,925 & 0/30 & 13.7 \\
Full RAR & 301,518 & --- & --- & --- & 9.3 \\
\bottomrule
\end{tabular}

\end{table}

The blatant stream is a weaker test of H2. The pre-filter, which is the layer
the blatant stream was built to exercise, quarantines its corruption before
it reaches the detector. The canary-and-rollback layer that H2 concerns is
therefore barely engaged. We keep the stream as a diagnostic. On the blatant
Airlines stream, Full RAR is indistinguishable from No-Response ($+549$,
unresolved) and behind Binary ($-22{,}796$, resolved), Always-Retrain and
Rollback-Only. The whole utility gap comes from lower predictive accuracy,
not from the costs of retraining, alerting, or rollback. Two compounding
causes account for it. The first is a weakness of Full RAR's own design: the
Gate-2 sample penalty of Section~\ref{sec:framework}. On
this stream Full RAR's challengers train on 1{,}890--2{,}289 labeled rows,
against the 2{,}700 available to ungated policies. Corruption is not concept
drift --- the relation the model should learn never moves --- so
onset-anchored purity buys nothing, while the 30\% sample deficit costs a
sample-hungry learner heavily. The second cause is the base learner. With the
same policy, data, and windows, a naive-Bayes learner scores 0.5833 where the
Hoeffding tree scores 0.5433. The tree fragments on a 6{,}585-valued
categorical (\texttt{Flight}) that recurs $\approx$80 times per value and
therefore meets no defensible identifier-exclusion rule
(Section~\ref{sec:datasets}). The learner choice moves utility by 21{,}573
units on this stream, larger than any policy effect measured anywhere in the
matrix. This is why we report the learner sensitivity itself as a finding.
Both causes also operate on the subtle stream, which shares the Airlines base
and the tree.

The seventh policy, Cost-Aware Binary, was specified with four predictions,
whose priority over the pilot is not verifiable (Section~\ref{sec:design}).
At pilot scale (five seeds), predictions 1--3 held. The gate never binds as
$\rho \to 0$, its retrain count falls monotonically
(21 $\to$ 20 $\to$ 19 $\to$ 5 at $\rho = 1, 10, 100, 1000$), and it narrows
Full RAR's high-$\rho$ margin. Prediction 4 (that Full RAR's advantage
concentrates on the safety streams) failed on the corruption stream, where
Cost-Aware Binary wins at every $\rho$ tested. It held on
rollback-then-drift, where the crossover lies in $(10, 100)$ and Full RAR
leads by $+1{,}763$ at $\rho = 100$ and $+4{,}391$ at $\rho = 1000$. At full
scale (all 13 streams, four simulated price points, 30 seeds;
Table~\ref{tab:cab}), the relative standing of Full RAR and Cost-Aware Binary depends on the stream and the price, the pattern its design document had named as the favorable outcome. Because that document
already contained the pilot outcome, we read this as consistent with, not
confirmatory of, the prediction. At $\rho \le 10$, Cost-Aware Binary beats
Full RAR (resolved) on \numcabwinsone{} of 13 streams. At $\rho = 100$, Full
RAR is ahead on five streams: it overtakes Cost-Aware Binary on ELEC2,
semi-abrupt and rollback-then-drift and stays ahead on NOAA and
INSECTS-abrupt. These five mix feasibility assessments (Table~\ref{tab:feas}).
At $\rho = 1000$ Full RAR wins on five streams while still losing on seven
(resolved; ELEC2's deficit is unresolved). On Covertype and semi-gradual it
wins by tens of thousands of units; there the cost gate stops retraining
almost entirely, and the baseline's accuracy collapses to \numcabacccover{}
and \numcabaccgradual{}. Cost-aware gating buys efficiency where retraining
is cheap; the cost-blind gates with safety arms earn their keep where it is
expensive and drift is real. Two caveats travel with that sentence. First,
part of the high-$\rho$ margin is the baseline's frozen horizon
over-suppressing --- a limitation stated in its design, not evidence for the
arms. Second, the baseline's retrain count is not monotone in $\rho$ on
\numcabmonofail{} of 13 streams (prediction 2 fails at scale). This is because
$\hat{H}$ is the observed inter-detection interval, and retraining changes the detector's
history, so the gate's threshold moves with the decisions it made. We left
$\hat{H}$ as specified and report the path dependence as the baseline's
property. The path dependence is also why the gate can produce \emph{more}
retrains than Binary on four of 52 cells. On Airlines at $\rho = 1$ the two
policies share an identical path of 27 detections and retrains until the
gate delays the 28th by 210 rows. After that, the staler served model
triggers 38 detections to Binary's 15, an excess of 23 that equals the
retrain excess to the unit.

\begin{table}[t]
\caption{Full RAR minus Cost-Aware Binary, paired by seed over 30 seeds, at
the four real price points ($^{*}$: resolved at two standard errors), and
the first price at which Full RAR is ahead with a resolved difference. Cost-Aware Binary is
not affine in $\rho$, so no analytic crossover is reported.}
\label{tab:cab}
\centering\scriptsize\setlength{\tabcolsep}{3pt}
\begin{tabular}{@{}lrrrrl@{}}
\toprule
Stream & $\rho{=}1$ & $\rho{=}10$ & $\rho{=}100$ & $\rho{=}1000$ & RAR ahead from \\
\midrule
ELEC2 & $-$77 & $-$323 & +1,882$^{*}$ & $-$516 & $\rho{=}100$ \\
Covertype & $-$32,812$^{*}$ & $-$29,956$^{*}$ & $-$11,886$^{*}$ & +56,945$^{*}$ & $\rho{=}1000$ \\
Poker & $-$0.1$^{*}$ & $-$1$^{*}$ & $-$10$^{*}$ & $-$100$^{*}$ & never \\
Airlines & $-$22,022$^{*}$ & $-$28,679$^{*}$ & $-$19,974$^{*}$ & $-$19,897$^{*}$ & never \\
NOAA & +6$^{*}$ & +48$^{*}$ & +489$^{*}$ & $-$2,300$^{*}$ & $\rho{=}1$ \\
INS-abrupt & +2,769$^{*}$ & +2,786$^{*}$ & +2,956$^{*}$ & +9,991$^{*}$ & $\rho{=}1$ \\
INS-gradual & $-$8,818$^{*}$ & $-$8,849$^{*}$ & $-$9,164$^{*}$ & $-$11,838$^{*}$ & never \\
INS-recurr. & $-$19,094$^{*}$ & $-$19,132$^{*}$ & $-$23,876$^{*}$ & $-$25,072$^{*}$ & never \\
S-Abrupt & $-$2,315$^{*}$ & $-$1,375$^{*}$ & +582$^{*}$ & +4,072$^{*}$ & $\rho{=}100$ \\
S-Gradual & $-$34,424$^{*}$ & $-$31,961$^{*}$ & $-$15,699$^{*}$ & +144,096$^{*}$ & $\rho{=}1000$ \\
S-Recurrent & $-$314$^{*}$ & $-$300$^{*}$ & $-$146$^{*}$ & $-$1,686$^{*}$ & never \\
S-Corrupt & $-$20,616$^{*}$ & $-$22,542$^{*}$ & $-$21,364$^{*}$ & $-$25,376$^{*}$ & never \\
S-RB-Drift & $-$304$^{*}$ & $-$420$^{*}$ & +1,775$^{*}$ & +4,413$^{*}$ & $\rho{=}100$ \\
\bottomrule
\end{tabular}

\end{table}

\subsection{H3 --- Alert Under Label Delay}
\label{sec:h3results}
The base matrix is run at the frozen $\tau = 800$. The label-delay axis is
carried by the ablation (Section~\ref{sec:h5results}). There, reducing delay
to $\tau = 200$ and $\tau = 0$ raises Full RAR's pooled utility on the Phase A streams by $+2{,}447$ and $+1{,}796$ (both resolved at ten seeds); delay has a measurable cost on these streams, with the INSECTS half of that
comparison provisional. At ten seeds on INSECTS the sign reverses:
$\tau = 0$ is $-\,21{,}179$ resolved against the $\tau = 800$ baseline
(Table~\ref{tab:abl}), so the delay effect is itself a stream property. Of
the two specified H3 quantities, the alert-to-retrain escalation delay is
\emph{not estimable} from the frozen ledger, which records alert counts, not
alert times. We report it as such rather than substitute a proxy. The
false-retrain count is computed from the frozen retrain indices on the five
ground-truth streams (Table~\ref{tab:h3}), with a strict boundary at the
first injected event. We read it as an \emph{upper bound}: the base streams
carry real drift before the injected onset, so a pre-onset retrain is not
necessarily false.

H3 has two conjuncts. On the false-retrain conjunct, Full RAR makes fewer pre-onset retrains than Binary on all five streams and has a lower pre-onset fraction on three.
On semi-abrupt, Full RAR retrains before the onset \numhthreerarpre{} times
across 30 seeds against Binary's \numhthreebinpre{}. That is a pre-onset
fraction of \numhthreerarfrac{} against \numhthreebinfrac{}. The fraction
ordering also holds on the recurrent and rollback-then-drift streams, but not
on semi-gradual. The utility conjunct (``at equal or better net utility'')
holds only on rollback-then-drift. On the other four ground-truth streams,
Full RAR is behind Binary at $\rho = 1$ (Table~\ref{tab:paired}). H3 is
therefore supported in its false-retrain component and not as specified. The
reduction is also not attributable to the alert arm alone, since Gates 2--3
are what withhold the retrain and the alert is what fires in its place. Two
rows are read with care. The semi-corrupt row is 1.0 for every
detector-triggered policy because the injected corruption sits in the
stream's tail, so it does not discriminate. Always-Retrain's fraction tracks
onset$/T$ on every stream because its schedule is set by stream length alone.
That is the metric validating itself rather than a policy result.

\begin{table}[t]
\caption{Retrains before the first injected event on the five ground-truth
streams: per-run mean count over 30 seeds (pre) and pooled pre-onset
fraction (frac). An upper bound on false retrains; see text.}
\label{tab:h3}
\centering\scriptsize\setlength{\tabcolsep}{3pt}
\begin{tabular}{@{}lrrrrrrrrr@{}}
\toprule
Stream & Onset & \multicolumn{2}{c}{Always-Retr.} & \multicolumn{2}{c}{Binary} & \multicolumn{2}{c}{Rollback-Only} & \multicolumn{2}{c}{Full RAR} \\
 & & pre & frac & pre & frac & pre & frac & pre & frac \\
\midrule
S-Abrupt & 28,442 & 10.0 & 0.62 & 17.0 & 0.68 & 4.8 & 0.56 & 2.0 & 0.46 \\
S-Gradual & 364,701 & 134.0 & 0.62 & 158.0 & 0.55 & 66.9 & 0.62 & 36.3 & 0.61 \\
S-Recurrent & 11,398 & 4.0 & 0.67 & 3.0 & 0.60 & 1.8 & 0.56 & 1.0 & 0.35 \\
S-Corrupt & 338,570 & 125.0 & 0.63 & 42.0 & 1.00 & 20.9 & 1.00 & 13.1 & 1.00 \\
S-RB-Drift & 28,442 & 10.0 & 0.62 & 17.0 & 0.74 & 4.8 & 0.70 & 2.0 & 0.40 \\
\bottomrule
\end{tabular}

\end{table}

\subsection{H4 --- Drift Type $\times$ Arm Interaction}
\label{sec:h4results}
H4 is specified on arm firing frequencies. Table~\ref{tab:h4arm} gives
Full RAR's arm rates per 100k instances by drift type. The three predictions
fare as follows. Abrupt drift was predicted to raise the retrain rate; it has
the lowest rate of the three drift-type groups (\hfourabruptretrains{} against
\hfourgradualretrains{} gradual and \hfourrecurrentretrains{} recurrent), so
the prediction is refuted. The predicted rise in the rollback rate under
gradual drift is not supported either: that rate is essentially equal to
abrupt (\hfourgradualrollbacks{} against \hfourabruptrollbacks{}) and below
recurrent (\hfourrecurrentrollbacks{}). Recurrent drift was predicted to
raise the alert rate. It has the highest of the three
(\hfourrecurrentalerts{}), but only marginally above gradual
(\hfourgradualalerts{}). With two streams per group and no test, this
prediction holds in direction only. Among the three drift types, recurrent
drift has the highest rate for all three arms, which is what a policy
re-triggered by returning concepts would show.

\begin{table}[t]
\caption{H4 on its specified quantity: Full RAR arm firings per 100k
instances by drift-type group (mean over streams and seeds). Descriptive
only; groups hold one to three streams.}
\label{tab:h4arm}
\centering\footnotesize\setlength{\tabcolsep}{3pt}
\begin{tabular}{@{}lrrrr@{}}
\toprule
Stream group & Streams & Retrains & Alerts & Rollbacks \\
\midrule
abrupt & 2 & 5.8 & 24.7 & 2.2 \\
gradual & 2 & 6.6 & 46.9 & 2.3 \\
recurrent & 2 & 9.2 & 48.6 & 3.3 \\
feature corruption & 1 & 2.4 & 48.7 & 1.2 \\
rollback then drift & 1 & 11.0 & 38.3 & 4.4 \\
real, type unspecified & 3 & 7.9 & 34.7 & 3.4 \\
low-information controls & 2 & 5.5 & 16.6 & 0.0 \\
\bottomrule
\end{tabular}

\end{table}

The policy-level picture differs. It was not part of specified H4, but the
decision map needs it. On abrupt drift the retrain-heavy policies are best.
Always-Retrain and Binary are the strongest policies on every abrupt stream,
and Full RAR is competitive only where its cycle fits between changes
(INSECTS-abrupt). On gradual drift, Rollback-Only beats Full RAR on
INSECTS-gradual and semi-gradual (Table~\ref{tab:paired}), though it does not
beat the retrain-heavy baselines. That difference combines Gates 2--3, the
alert arm, and the typed holds (Section~\ref{sec:policies}). On recurrent
drift, Full RAR is behind Binary on INSECTS-recurrent at every $\rho$, and on
the semi-recurrent stream it falls below No-Response on 25 of 30 seeds.
Always-Retrain, refreshing every $W_{\text{train}}$ instances, tracks
returning concepts best. Recurrent drift rewards cheap, frequent refits, and
the decision map reflects that. Table~\ref{tab:dmap} is the decision map.
Within each drift-type group, policies are ranked inside each stream and the
ranks averaged, so long streams cannot dominate on raw utility. A bold entry
is a winner that also beats its nearest rival under the paired rule on every
stream in the group; unmarked entries are descriptive only.

\begin{table}[t]
\caption{Decision map v1 (drift type $\times$ $\rho$ $\to$ best policy by
mean within-stream rank). Bold: resolved and consistent across the group;
$^{=}$: tied rank winners; $^{h}$: heterogeneous across the group's streams.
AR Always-Retrain, Bin Binary, RB Rollback-Only, RAR Full RAR, NR
No-Response.}
\label{tab:dmap}
\centering\footnotesize\setlength{\tabcolsep}{3pt}
\begin{tabular}{@{}lccccc@{}}
\toprule
Stream group & $\rho=1$ & $\rho=5$ & $\rho=10$ & $\rho=100$ & $\rho=200$ \\
\midrule
abrupt & \textbf{AR} & \textbf{AR} & \textbf{AR} & RAR$^{h}$ & \textbf{RAR} \\
gradual & AR/Bin$^{=}$ & AR/Bin$^{=}$ & AR/Bin$^{=}$ & \textbf{AR} & AR/RB$^{=}$ \\
recurrent & AR/Bin$^{=}$ & AR/Bin$^{=}$ & AR/Bin$^{=}$ & Bin$^{h}$ & Bin$^{h}$ \\
feature corruption & \textbf{Bin} & \textbf{Bin} & \textbf{Bin} & \textbf{Bin} & \textbf{Bin} \\
rollback then drift & \textbf{AR} & \textbf{AR} & \textbf{AR} & \textbf{RAR} & \textbf{RAR} \\
natural drift unspecified & AR/Bin$^{=}$ & AR$^{h}$ & AR$^{h}$ & Bin$^{h}$ & RB$^{h}$ \\
low information controls & RB$^{h}$ & RB$^{h}$ & RB$^{h}$ & NR$^{h}$ & NR$^{h}$ \\
\bottomrule
\end{tabular}

\end{table}

\subsection{H5 and Ablations}
\label{sec:h5results}
\textbf{Canary value (H5).} H5 is specified on the per-deployment cost of
bad retrains with and without the canary inside the same policy. Two
policy-level comparisons bear on it, and both run against it. The first is in
the base matrix. Rollback-Only is Binary plus the two-stage canary and
nothing else (Table~\ref{tab:h5canary}). At $\rho = 1$, adding the canary
lowers net utility on \hfivelose{} of 13 streams (resolved) and raises it on
\hfivewin{} (NOAA). Maximum drawdown is lower with the canary on
\hfiveddlower{} of 13. On these streams the canary rejects many challengers
that would have paid, up to 57 rollbacks per run on the Covertype streams.
Its protection against bad deployments does not offset the retrains it
discards.

The second is the registered follow-up E2a (Table~\ref{tab:e2a}), which
switches the canary off inside Full RAR and changes nothing else. Removing it
raises net utility on \etwolose{} of 13 streams (resolved) and lowers it on
\etwowin{} (semi-recurrent, by 286); Poker is unchanged. Every rollback in RAR
fires from a canary state, so the no-canary policy also never rolls back.
E2a removes the canary and the rollback arm together. The
no-canary policy retrains more often (99 against 60 retrains per run on the
Covertype streams), and at $\rho = 1$ on these streams the extra retrains
pay. Because the no-canary policy retrains more, the sign can reverse at high $\rho$. The two
trajectories diverge after the first retrain, so E2a is a policy-level
ablation, not the specified per-deployment quantity.

That quantity was to be measured by E2b, which scores the incumbent and each
challenger on the same 5{,}000 instances after the challenger is created.
The pre-launch check found that 167 of 5{,}850 creation events have fewer
than 5{,}000 later instances. These events are concentrated on the short
streams (30 of 60 on NOAA), and the registered definition gave no rule for
them. Excluding them would bias the short streams, so E2b was stopped
rather than patched with a censoring rule chosen after the problem was seen.
H5 as specified remains untested. E4, the alert-arm removal, was stopped
because in the current loop the policy's alert flag suppresses only the alert
charge, while the alert monitoring clock still runs. A faithful removal would
change the decision loop, and we did not run it.

\begin{table}[t]
\caption{The canary as a one-factor change: Rollback-Only minus Binary,
paired by seed, at $\rho = 1$ ($^{*}$ resolved), with Rollback-Only's mean
rollbacks per run and both policies' mean maximum accuracy drawdown.}
\label{tab:h5canary}
\centering\scriptsize\setlength{\tabcolsep}{3pt}
\begin{tabular}{@{}lrrrrr@{}}
\toprule
Stream & $\Delta U(1)$ & 2$\cdot$SEM & Rollbacks/run & Drawdown RB & Drawdown Bin \\
\midrule
ELEC2 & $-$246$^{*}$ & 126 & 6.8 & 0.572 & 0.600 \\
Covertype & $-$18,595$^{*}$ & 1,314 & 57.1 & 0.967 & 0.958 \\
Poker & 0 & 0 & 0.0 & 0.174 & 0.174 \\
Airlines & $-$9,530$^{*}$ & 1,351 & 13.7 & 0.626 & 0.636 \\
NOAA & +59$^{*}$ & 35 & 1.5 & 0.249 & 0.324 \\
INS-abrupt & $-$12,383$^{*}$ & 1,667 & 2.6 & 0.629 & 0.610 \\
INS-gradual & $-$8,471$^{*}$ & 4 & 1.0 & 0.676 & 0.632 \\
INS-recurr. & $-$29,635$^{*}$ & 10,335 & 10.3 & 0.705 & 0.626 \\
S-Abrupt & $-$2,094$^{*}$ & 182 & 5.2 & 0.705 & 0.720 \\
S-Gradual & $-$15,764$^{*}$ & 1,128 & 56.7 & 0.953 & 0.960 \\
S-Recurrent & $-$254$^{*}$ & 13 & 1.8 & 0.567 & 0.560 \\
S-Corrupt & $-$8,634$^{*}$ & 1,118 & 9.4 & 0.609 & 0.636 \\
S-RB-Drift & $-$3,006$^{*}$ & 304 & 4.7 & 0.782 & 0.650 \\
\bottomrule
\end{tabular}

\end{table}

\begin{table}[t]
\caption{E2a: Full RAR minus Full RAR without the canary, paired by seed,
$\rho = 1$ ($^{*}$ resolved), with mean retrains per run and the no-canary
policy's rollbacks per run (zero everywhere: rollback fires only from canary
states).}
\label{tab:e2a}
\centering\scriptsize\setlength{\tabcolsep}{3pt}
\begin{tabular}{@{}lrrrrrr@{}}
\toprule
Stream & $\Delta U(1)$ & 2$\cdot$SEM & $+$/$-$ & Retr.\ RAR & Retr.\ no-can. & Rollb.\ no-can. \\
\midrule
ELEC2 & $-$1,618$^{*}$ & 450 & 0/30 & 4.3 & 9.0 & 0.0 \\
Covertype & $-$14,242$^{*}$ & 1,744 & 0/30 & 60.6 & 99.0 & 0.0 \\
Poker & 0 & 0 & 0/0 & 0.0 & 0.0 & 0.0 \\
Airlines & $-$17,937$^{*}$ & 1,572 & 0/30 & 20.0 & 24.0 & 0.0 \\
NOAA & $-$167$^{*}$ & 0 & 0/30 & 2.0 & 2.0 & 0.0 \\
INS-abrupt & $-$11,854$^{*}$ & 1,790 & 0/30 & 6.9 & 11.0 & 0.0 \\
INS-gradual & $-$14,934$^{*}$ & 15 & 0/30 & 4.0 & 5.0 & 0.0 \\
INS-recurr. & $-$5,528$^{*}$ & 5,363 & 7/23 & 11.8 & 16.0 & 0.0 \\
S-Abrupt & $-$1,376$^{*}$ & 224 & 0/30 & 4.3 & 8.0 & 0.0 \\
S-Gradual & $-$18,621$^{*}$ & 1,739 & 0/30 & 59.9 & 99.0 & 0.0 \\
S-Recurrent & +286$^{*}$ & 5 & 30/0 & 2.9 & 3.0 & 0.0 \\
S-Corrupt & $-$14,876$^{*}$ & 837 & 0/30 & 13.1 & 23.0 & 0.0 \\
S-RB-Drift & $-$1,620$^{*}$ & 6 & 0/30 & 5.0 & 9.0 & 0.0 \\
\bottomrule
\end{tabular}

\end{table}

\begin{table}[t]
\caption{One-factor ablation of Full RAR, pooled paired $\Delta U(1)$ versus
the ratified baseline. Phase A: ELEC2, NOAA, semi-corrupt, and rollback-then-drift, ten seeds. Phase B: INSECTS-abrupt and
-recurrent, ten seeds. R: resolved at two standard errors; n.r.: not resolved;
$^{*}$: sign flip resolved in both phases. Phase-B cells on which the two
INSECTS streams disagree in sign are flagged in the artifact's per-stream
tables.}
\label{tab:abl}
\centering\scriptsize\setlength{\tabcolsep}{3pt}
\begin{tabular}{@{}lrlrlc@{}}
\toprule
Cell & Phase A & & Phase B & & Cross-phase \\
\midrule
canary.commit\_z=0.0 & +884 & n.r. & 0 & n.r. & no sign \\
canary.delta=0.01 & +853 & n.r. & +4,548 & R & same \\
canary.delta=0.05 & +30 & n.r. & $-$263 & n.r. & flip \\
canary.p1=0.1 & +2,406 & R & $-$11,456 & R & flip$^{*}$ \\
canary.p1=0.4 & +996 & n.r. & $-$10,402 & R & flip \\
canary.routing=iid & +1,430 & R & +2,678 & n.r. & same \\
canary.w1=150, canary.w2=300 & +2,229 & R & +5,232 & n.r. & same \\
canary.w1=600, canary.w2=1200 & +2,055 & R & +5,369 & n.r. & same \\
detector.warmup=30 & +1,420 & n.r. & $-$3,943 & n.r. & flip \\
gates.guard\_gap\_g=0 & $-$1,361 & R & $-$10,511 & n.r. & same \\
gates.guard\_gap\_g=100 & +2,059 & R & +10,512 & R & same \\
gates.guard\_gap\_g=400 & +2,852 & R & $-$8,113 & n.r. & flip \\
gates.theta\_L=0.5 & +553 & n.r. & +98 & n.r. & same \\
gates.theta\_R=0.03 & +1,986 & R & +6,109 & n.r. & same \\
gates.theta\_R=0.08 & +1,766 & R & $-$47,563 & R & flip$^{*}$ \\
holds.delta\_crit\_mult=2.0 & +458 & R & $-$886 & n.r. & flip \\
prefilter.enabled=false & +219 & n.r. & +6,515 & R & same \\
prefilter.min\_continuous\_card=2 & $-$63 & n.r. & 0 & n.r. & no sign \\
stream.label\_delay\_tau=0 & +1,796 & R & $-$21,179 & R & flip$^{*}$ \\
stream.label\_delay\_tau=200 & +2,447 & R & $-$5,102 & n.r. & flip \\
windows.w\_train=4000 & +3,120 & R & $-$43,507 & R & flip$^{*}$ \\
\bottomrule
\end{tabular}

\end{table}

\begin{table}[t]
\caption{The sharpest heterogeneity case, inside Phase A alone:
\texttt{canary.p1 = 0.1} per stream, same ten seeds.}
\label{tab:p1}
\centering\footnotesize
\begin{tabular}{@{}lrrl@{}}
\toprule
Stream & Mean $\Delta U$ & $2\cdot$SEM & Verdict \\
\midrule
semi\_corrupt\_airlines & +12,384 & 2,005 & RESOLVED \\
semi\_rollback\_then\_drift & $-$2,945 & 16 & RESOLVED \\
river\_elec2 & +191 & 771 & not resolved \\
noaa\_weather & $-$4.7 & 4.8 & not resolved \\
\bottomrule
\end{tabular}

\end{table}

\textbf{Parameter effects are properties of the stream, not of the method.}
Table~\ref{tab:abl} reports the 21 one-factor cells run in both phases. The
two H6-only $\kappa_{\text{long}}$ cells are reported in
Section~\ref{sec:h6results}. Of the \numablcommon{} cells comparable across
the two phases at ten seeds each, \numablflips{} flip sign between the
cheap-stream group and INSECTS, and \numablbothres{} are resolved in both
phases with opposite signs. Within Phase B alone, \numablbconf{} of 21 cells
have opposite signs on its two streams. The $\theta_L = 0.5$ cell is
\numthlabr{} on INSECTS-abrupt and $+$\numthlrec{} on INSECTS-recurrent;
its pooled mean of $+98$ is produced by two large effects in opposite
directions. The sharpest case does not depend on the phase-B seed count at
all (Table~\ref{tab:p1}). Routing 10\% rather than 20\% of traffic to the
challenger is worth $+12{,}384$ on the corruption stream and $-2{,}945$ on
rollback-then-drift, both resolved, on identical seeds. Its pooled mean of $+2{,}406$ hides two resolved effects of opposite sign.
Consequently, every ablation table in the released artifact is per-stream by
construction, and pooled rows carry a sign-conflict flag.

Four cells deserve individual mention. Guard gap $g = 0$ is the only
\emph{negative} resolved cell in Phase A ($-1{,}361$). Removing the guard
hurts, which confirms in utility the contamination-bias evidence of
Section~\ref{sec:erroranalysis}. $W_{\text{train}} = 4{,}000$ is $+3{,}120$
resolved on the cheap streams, the direction the Gate-2 sample-penalty
analysis of Section~\ref{sec:framework} predicts. It is \numwtrainb{}
resolved on INSECTS, where the larger anchored window spans more of the
stream's own change; the sample-penalty trade reverses on long, frequently
changing streams. For $\kappa_{\text{long}} \in \{2500, 10000\}$, both cells
measure a difference of exactly $0.0$. The axis ran, but the hypothesis it
serves was not tested, because on the frozen stream the hold that spans the
drift is opened by a quality rollback and is therefore governed by
$\kappa_{\text{short}}$; $\kappa_{\text{long}}$ is never read
(Section~\ref{sec:h6results}). The pre-filter-removal cell is unresolved on
the cheap streams ($+219$) and splits on INSECTS. It is exactly \numpfabr{}
on every seed of INSECTS-abrupt, where nothing is ever quarantined, and
$+$\numpfrec{} resolved on INSECTS-recurrent, where quarantining instances
from detector updates plausibly delays detections the policy needs. The
pre-filter's justification was never net utility. It was the
detector-poisoning failure the pre-filter prevents and the quarantine recall
it delivers. These cells put a measurement beside that claim and mark where
the two-layer defense has a cost.

\begin{table}[t]
\caption{Stage-1 routing: stratified (matrix default) versus i.i.d.\ at 30
seeds, paired by seed. Per-routing columns give mean $\pm$ s.d.\ over seeds; the paired difference gives mean $\pm$ SEM.}
\label{tab:iid}
\centering\footnotesize\setlength{\tabcolsep}{3pt}
\begin{tabular}{@{}llrrrl@{}}
\toprule
Stream & Policy & Stratified & i.i.d. & Paired $\Delta$ & Verdict \\
\midrule
ELEC2 & Full RAR & 32,623 $\pm$ 1,233 & 32,952 $\pm$ 1,050 & $-$329 $\pm$ 289 & n.r. \\
ELEC2 & Rollback-Only & 32,292 $\pm$ 345 & 32,123 $\pm$ 342 & +169 $\pm$ 92 & n.r. \\
INS-abrupt & Full RAR & 189,793 $\pm$ 4,902 & 183,181 $\pm$ 2,070 & +6,612 $\pm$ 1,037 & resolved \\
INS-abrupt & Rollback-Only & 174,641 $\pm$ 4,566 & 169,622 $\pm$ 8,959 & +5,019 $\pm$ 1,734 & resolved \\
S-RB-Drift & Full RAR & 31,300 $\pm$ 16 & 30,310 $\pm$ 1,353 & +990 $\pm$ 247 & resolved \\
S-RB-Drift & Rollback-Only & 28,087 $\pm$ 832 & 27,928 $\pm$ 662 & +158 $\pm$ 188 & n.r. \\
\bottomrule
\end{tabular}

\end{table}

\textbf{Routing.} Stratified routing was adopted on first-principles grounds:
it gives an unbiased, lower-variance canary estimator, and it is
production-realistic. Its documented justification was subsequently retracted
when a 30-seed comparison on ELEC2 showed i.i.d.\ routing slightly higher and
less variable (Table~\ref{tab:iid}, unresolved). Measured directly, the
stratified estimator \emph{is} unbiased and 2.3$\times$ tighter (error s.d.\
0.0117 vs.\ 0.0274 over eleven 1{,}500-instance windows). Yet it does not
translate into higher utility there. A tighter estimate of the canary window
is not a better estimate of the future, and the more confident estimator
resolves marginal challengers as rollbacks on a stream where promotion pays.
On the other two streams measured, stratified routing is better ($+6{,}612$
on INSECTS-abrupt, $+990$ on rollback-then-drift, both resolved). The default
stands as specified; the estimator-versus-outcome distinction is developed in
Section~\ref{sec:limitations}.

\subsection{H6 --- Post-Rollback Recovery}
\label{sec:h6results}
\begin{table}[t]
\caption{H6 designated stream, two-tier (as designed) versus single-tier
cooldown, 30 paired seeds. Primary metric --- fraction of seeds hold-locked
when the genuine drift arrives --- is $0/30$ for both variants. Per-variant columns give mean $\pm$ s.d.\ over seeds; the paired difference gives mean $\pm$ SEM.}
\label{tab:h6}
\centering\footnotesize\setlength{\tabcolsep}{3pt}
\begin{tabular}{@{}lrrrl@{}}
\toprule
Metric & Two-tier & Single-tier & Paired $\Delta$ & Verdict \\
\midrule
Accuracy & 0.6909 $\pm$ 0.0003 & 0.6909 $\pm$ 0.0004 & -0.0000 $\pm$ 0.0000 & n.r. \\
Net utility ($\rho=1$) & 31,300.4 $\pm$ 15.6 & 31,302.7 $\pm$ 17.8 & -2.3 $\pm$ 2.1 & n.r. \\
Retrains & 5.00 $\pm$ 0.00 & 4.00 $\pm$ 0.00 & +1.00 $\pm$ 0.00 & resolved \\
Alerts & 17.37 $\pm$ 0.49 & 16.00 $\pm$ 0.00 & +1.37 $\pm$ 0.09 & resolved \\
Rollbacks & 2.00 $\pm$ 0.00 & 1.00 $\pm$ 0.00 & +1.00 $\pm$ 0.00 & resolved \\
Commits & 2.00 $\pm$ 0.00 & 2.00 $\pm$ 0.00 & +0.00 $\pm$ 0.00 & n.r. \\
Max drawdown & 0.7489 $\pm$ 0.0072 & 0.7483 $\pm$ 0.0071 & +0.0007 $\pm$ 0.0017 & n.r. \\
Time below no-skill & 9,023.07 $\pm$ 42.31 & 9,023.17 $\pm$ 43.28 & -0.10 $\pm$ 0.83 & n.r. \\
\bottomrule
\end{tabular}

\end{table}

\textbf{The specified discriminator was inactive, and the cause is
identified.} Neither variant was hold-locked when the genuine drift arrived on
any of 30 seeds (Table~\ref{tab:h6}). The primary metric, first committed
with the first H6 payloads (Section~\ref{sec:design}), therefore returns
null, and we report it as such. A direct probe, not code reading, verified
that the single-tier variant implements the specified comparison. It applies
$\kappa_{\text{long}}$ (5{,}000 ticks against the two-tier's 1{,}500),
retains no $q_t$ early exit (releases at tick 5{,}000 where the hard hold
releases at tick 1), and admits no override. What defeats the scenario is
upstream of the mechanism. On every seed, an unrelated R1 quality rollback
fires $\approx$4{,}000 rows \emph{before} the corruption window. The two-tier
system takes its short cooldown and resumes with a canary in flight when
corruption begins. It then takes the R2 rollback into a hard hold and is
released by $q_t$ 150 rows before the genuine drift, which is the design
doing its job. The single-tier system is still inside its 5{,}000-row
cooldown when corruption begins. Having no canary in flight, it cannot take
the corruption rollback at all (one rollback per run against two). Its hold
expires with 940 corrupt rows remaining, too few to detect, retrain, and
route before the feed recovers. The naive cooldown thus locks through the
\emph{corruption} and thereby avoids the very rollback that was to lock it
through the \emph{drift}. This is a richer failure mode than hypothesized:
the cooldown holds at the wrong time rather than merely too long. What
survives is descriptive and seed-stable. Single-tier spends 11.0\% of the
stream hold-locked against two-tier's 6.3--6.9\% --- about 1.6--1.7$\times$ as long in hold --- for a utility difference of $-2.3 \pm 2.1$, unresolved. That extra hold time shows no resolved utility effect, and it is not the specified discriminator. Two
consequences follow. The $\kappa_{\text{long}}$ ablation is mechanically
inert on this stream (Section~\ref{sec:h5results}). The
$\kappa_{\text{long}} = 10{,}000$ cell re-engages no lock on this stream
either (difference exactly 0.0, Section~\ref{sec:h5results}). A stream on
which it did would act through the quality rollback and would test a broader
undifferentiated-cooldown claim rather than specified H6.

\textbf{The post-hoc follow-up} (Table~\ref{tab:h6fu}) was designed after
this null and is reported as such, never as the prospectively specified
result. A new stream places 2{,}266 rows of feature corruption first, then a
500-row clean gap, then the genuine label drift, so that the corruption
rollback opens the first hold. A seed-0 preflight verified all four
pre-launch mechanism conditions before the 30-seed run. The stream passed the
same validation trio as every other injected stream, with $q_t$ recall 0.959
and precision 0.934 against its corruption span at a 0.38\% clean
false-positive rate. The discriminator activates. Single-tier's 5{,}000-row
cooldown spans the drift on 30/30 seeds, and two-tier's hard hold is released
by $q_t$ 146 rows before it on 30/30. Under that condition two-tier is
$+0.045$ in accuracy and $+2{,}027$ in utility, resolved, with utility
seed-dispersion of 14 against 1{,}531. That dispersion is structured. Four of
the single-tier seeds form a degraded-recovery subgroup, with two rollbacks
and two commits where the other 26 show one and three. This subgroup is
separated from the rest by 0.093 in accuracy, so the naive cooldown
does not merely lose on average; on some seeds it fails to recover at all.
Because the stream was constructed to activate the mechanism, the margin is a
property of the construction: move the drift 200 rows earlier and both
variants are locked. This is therefore an existence proof that two-tier
release can respond where a fixed cooldown cannot, not an effect-size
estimate. Together the two results show that the advantage is real but
depends on when the rollback falls relative to the drift. The frozen stream
shows that this timing does not always occur.

\begin{table}[t]
\caption{Post-hoc H6 follow-up stream, two-tier versus naive single-tier, 30
paired seeds, regenerated from the per-cell payloads. $^{*}$ denotes a
difference resolved under the routing-variation rule; because the stream was
constructed after the H6 null to activate the mechanism, these values are
descriptive rather than confirmatory. The mechanism metric for this follow-up
is the first row. The drawdown difference is resolved but present on only
\hsixddseeds{} of 30 seeds.}
\label{tab:h6fu}
\centering\scriptsize\setlength{\tabcolsep}{2.5pt}
\begin{tabular}{@{}lrrrrl@{}}
\toprule
Measure & Two-tier & Single-tier & Paired $\Delta$ & 2$\cdot$SEM & Seeds $+$/$-$/0 \\
\midrule
Locked through genuine drift & 0/30 & 30/30 & --- & --- & primary \\
Accuracy & 0.6695 & 0.6247 & +0.0448$^{*}$ & 0.0123 & 30/0/0 \\
Net utility at $\rho = 1$ & 30,329 & 28,301 & +2,027$^{*}$ & 558 & 30/0/0 \\
Retrains & 5.0 & 4.0 & +1.0$^{*}$ & 0.0 & 30/0/0 \\
Time below no-skill & 13,313 & 16,391 & $-$3,078$^{*}$ & 988 & 0/30/0 \\
Maximum accuracy drawdown & 0.794 & 0.808 & $-$0.014$^{*}$ & 0.013 & 0/4/26 \\
\bottomrule
\end{tabular}

\end{table}

\subsection{Tail Metrics and the Negative Control}
\label{sec:tailresults}
\begin{table}[t]
\caption{Tail metrics over 30 seeds: worst-seed net utility at $\rho = 1$ and
mean maximum rolling-accuracy drawdown (500-instance windows).}
\label{tab:tail}
\centering\scriptsize\setlength{\tabcolsep}{1.6pt}
\begin{tabular}{@{}lrrrrrrrr@{}}
\toprule
Stream & \multicolumn{2}{c}{Full RAR} & \multicolumn{2}{c}{Binary} & \multicolumn{2}{c}{Always-Retr.} & \multicolumn{2}{c}{Rollback-Only} \\
 & worst $U$ & drawdown & worst $U$ & drawdown & worst $U$ & drawdown & worst $U$ & drawdown \\
\midrule
ELEC2 & 30,898 & 0.572 & 32,538 & 0.600 & 33,157 & 0.654 & 31,461 & 0.572 \\
Covertype & 362,827 & 0.889 & 401,208 & 0.958 & 391,981 & 0.988 & 375,214 & 0.967 \\
Poker & 513,697 & 0.174 & 513,697 & 0.174 & 513,318 & 0.174 & 513,697 & 0.174 \\
Airlines & 297,253 & 0.641 & 330,968 & 0.636 & 324,869 & 0.752 & 315,458 & 0.626 \\
NOAA & 12,452 & 0.250 & 12,446 & 0.324 & 12,585 & 0.304 & 12,428 & 0.249 \\
INS-abrupt & 182,776 & 0.641 & 187,024 & 0.610 & 194,122 & 0.818 & 164,616 & 0.629 \\
INS-gradual & 64,218 & 0.614 & 73,128 & 0.632 & 82,919 & 0.584 & 64,634 & 0.676 \\
INS-recurr. & 189,388 & 0.681 & 219,997 & 0.626 & 245,533 & 0.710 & 154,930 & 0.705 \\
S-Abrupt & 29,419 & 0.781 & 31,093 & 0.720 & 32,335 & 0.660 & 28,421 & 0.705 \\
S-Gradual & 352,379 & 0.909 & 392,852 & 0.960 & 386,761 & 0.988 & 369,506 & 0.953 \\
S-Recurrent & 10,970 & 0.560 & 11,299 & 0.560 & 11,180 & 0.570 & 10,986 & 0.567 \\
S-Corrupt & 289,037 & 0.639 & 314,664 & 0.636 & 313,040 & 0.752 & 300,389 & 0.609 \\
S-RB-Drift & 31,266 & 0.749 & 31,093 & 0.650 & 31,687 & 0.656 & 27,295 & 0.782 \\
\bottomrule
\end{tabular}

\end{table}
Table~\ref{tab:tail} reports the prospectively specified catastrophe-avoidance
metrics. Worst-seed utility tracks mean utility closely for all policies; the
ordering of Table~\ref{tab:mainresults} is not an artifact of a few good
seeds. Maximum drawdown is governed by the stream far more than by the
policy. The negative control behaves exactly as required. On Poker, with no
drift, Full RAR fires zero retrains and zero rollbacks on every seed. Its
utility deficit against No-Response is exactly \numpokergap{}, the price of
two soft-anomaly alerts, and Always-Retrain pays 379 retrains for no accuracy
gain. On the rollback-then-drift stream, Full RAR holds accuracy
seed-dispersion at \numrbdsd{} against Rollback-Only's \numrbdrosd{}. This
more than fifty-fold stability difference is invisible in a table of means.

\subsection{Second-Learner Robustness}
\label{sec:e3}
All policy comparisons above use one Hoeffding tree, and the Airlines
diagnostic shows the learner can move utility more than any policy. E3
repeats No-Response, Always-Retrain, Binary and Full RAR with a Gaussian
naive-Bayes learner on five streams (Table~\ref{tab:e3}). It carries no
hypothesis. The sign of Full RAR against Binary agrees with the tree, with
both differences resolved, on \ethreeagreeone{} of 5 streams at $\rho = 1$ and
\ethreeagreehund{} of 5 at $\rho = 100$. Two results do not transfer.
Rollback-then-drift is a FEASIBLE stream on which Full RAR beats Binary with
the tree, and it reverses with naive Bayes, so at least one FEASIBLE-stream
outcome in Table~\ref{tab:feas} depends on the learner. On
semi-gradual at $\rho = 100$, Full RAR moves ahead of Binary with naive Bayes
while staying behind with the tree. The INSECTS-abrupt win and the
INSECTS-recurrent and subtle-corruption losses hold under both learners.

\begin{table}[t]
\caption{E3: Full RAR minus Binary, paired by seed ($^{*}$ resolved), with
Gaussian naive Bayes (NB) and the Hoeffding tree (HT) of the base matrix
(E1 for the subtle stream).}
\label{tab:e3}
\centering\scriptsize\setlength{\tabcolsep}{3pt}
\begin{tabular}{@{}llrrrr@{}}
\toprule
 & & \multicolumn{2}{c}{$\rho = 1$} & \multicolumn{2}{c}{$\rho = 100$} \\
Stream & Feasibility & NB & HT & NB & HT \\
\midrule
INS-abrupt & FEASIBLE & +4,472$^{*}$ & +2,769$^{*}$ & +4,593$^{*}$ & +2,956$^{*}$ \\
INS-recurr. & INFEASIBLE & $-$38,970$^{*}$ & $-$19,094$^{*}$ & $-$38,967$^{*}$ & $-$19,510$^{*}$ \\
S-Gradual & MIXED & $-$16,863$^{*}$ & $-$28,020$^{*}$ & +6,309$^{*}$ & $-$7,372$^{*}$ \\
S-RB-Drift & FEASIBLE & $-$4,050$^{*}$ & +207$^{*}$ & $-$2,300$^{*}$ & +1,864$^{*}$ \\
S-Subtle & --- & $-$16,318$^{*}$ & $-$25,819$^{*}$ & $-$13,632$^{*}$ & $-$22,452$^{*}$ \\
\bottomrule
\end{tabular}

\end{table}

\subsection{Error Analysis}
\label{sec:erroranalysis}
\begin{table}[t]
\caption{Gate-3 input quality on the ground-truth streams (ADWIN,
estimated-onset anchor): absolute bias of $\hat{A}_{\text{pre}}$ against the
true pre-disturbance accuracy, with and without the guard gap.}
\label{tab:tl14}
\centering\footnotesize\setlength{\tabcolsep}{3pt}
\begin{tabular}{@{}llrrr@{}}
\toprule
Stream & Type & Onset lag & $|$bias$|$, $g{=}0$ & $|$bias$|$, $g{=}200$ \\
\midrule
S-Abrupt & abrupt & 191 & 0.081 & 0.003 \\
S-Corrupt & corrupt & 7935 & 0.231 & 0.222 \\
S-Gradual & gradual & 15 & 0.004 & 0.037 \\
S-Recurrent & recurrent & 767 & 0.423 & 0.313 \\
S-RB-Drift & rollback\_then\_drift & 639 & 0.208 & 0.151 \\
S-Subtle & corrupt & 3071 & 0.062 & 0.050 \\
\bottomrule
\end{tabular}

\end{table}
\textbf{Gate 3's accuracy varies sharply across drift types and helps explain
several outcomes} (Table~\ref{tab:tl14}). On abrupt drift the guarded
reference-window bias is 0.003, an order of magnitude below $\theta_R =
0.05$, so the reference-accuracy estimate there has low measured bias. The bias is not uniformly
low, however; it is 0.151 on the rollback-then-drift stream. On
the recurrent stream the residual is 0.313, six times $\theta_R$. With
concepts 2{,}253 rows apart, a reference window of 1{,}000 instances plus
guard spans previous concepts, and Gate 3 is effectively reading noise. This
helps explain the recurrent-stream failures of
Section~\ref{sec:h4results}. On gradual drift the guard \emph{hurts} (0.004
$\to$ 0.037), because the estimated-onset lag is only 15 rows and a 200-row
guard discards clean evidence. The fixed guard's limitation cuts both ways.

Five cases are recorded in the artifact. Each carries a post-hoc analysis run
identifier explicitly tagged as \emph{not} an original execution (matrix
tracking was disabled by design), a signature over its sixty source cells,
representative seeds, and effect sizes. The five are three matrix losses to
Binary (semi-corrupt Airlines, INSECTS-gradual, INSECTS-recurrent), one
mechanism diagnostic, and the specified H6 null. Three are traced here.
\emph{(i) Commits that changed nothing:} on NOAA, Full RAR spends 45.7\% of
an 18{,}159-row stream in canary states. The stream is barely 3.8
first-order adaptation cycles long (\noaacyclescorr{} with the canary label
waits of Eq.~(\ref{eq:cycle})). Full RAR commits two challengers trained on
1{,}927 and 2{,}225 labels whose served accuracy matches the frozen incumbent's to six decimal places; the ledger records the commits and no accuracy
change. \emph{(ii) Decision leverage, not timing:} Full RAR's ELEC2 utility
is trimodal across seeds (11/9/10), with accuracy s.d.\ 0.027 against
Rollback-Only's 0.008 under the same canary. Gate 2's fixed
$\theta_L W_{\text{train}}$ offset synchronizes episode timing across seeds,
so the dispersion comes from which marginal challengers a routing draw
promotes, each then serving thousands of instances. \emph{(iii) The learner
inversion} on Airlines (Section~\ref{sec:h2results}).

\section{Discussion: Implementing RAR in Practice}
\label{sec:discussion}
The state-machine formulation is deliberately chosen for implementability:
each state and transition maps onto commodity MLOps components.
Organizations with a model registry, a monitoring layer, and a serving layer
that supports traffic splitting can realize RAR by wiring existing
components. Gate 1 is wired to their drift monitor, Gate 2 to label-pipeline
lag metrics, and Gate 3 to their offline evaluation store. The rollback arm
is wired to registry version history, and the canary to their existing
staged-rollout mechanism. Adopters can start from the $(W_1, W_2, \delta)$
defaults tested in our sensitivity analysis, then evaluate them on their own streams. Research settings can use the released harness
directly. Organizations without model versioning or traffic splitting cannot
adopt the rollback arm cheaply. For them, RAR functions as an architectural
argument for which infrastructure to build first---versioning before
automation. We expect canary traffic splitting to be the hardest component in production; the rest of RAR is conditional logic over quantities that many mature ML platforms already compute.

Threshold-setting guidance follows from the ablations. First, $\theta_R$
should be set no lower than the noise floor of the rolling-accuracy
estimator at the chosen window. Second, $W_2$ should be sized from the
paired power bound of Appendix~\ref{app:power}, using the discordance the
operator observes between incumbent and challenger. Third,
$\kappa_{\text{long}}$ should exceed the expected duration of transient
corruption events in the operator's environment; the corruption-clear early
exit bounds the cost when faults resolve sooner. Finally,
$\Delta_{\text{crit}}$ should be set well above $\theta_R$, so that the
emergency override cannot immediately re-trigger a retrain that just failed
canary.

\section{Limitations}
\label{sec:limitations}
Six limitations bound our claims. (1) All streams are tabular classification streams; evaluation on vision, text, and regression streams is future work.
(2) The canary is a seeded-sampling simulation. It is a reproducible proxy
for live traffic splitting, not a measurement of it; ecological validation
requires production deployment. (3) Label delay is modeled as fixed $\tau$,
not adaptive or feedback-dependent. (4) Costs are stationary and known up to
$\rho$; estimating $\rho$ from operational telemetry is an open problem we do
not solve. (5) The framework maintains one challenger model;
multi-challenger rollouts and champion--challenger portfolios are natural
extensions. (6) Our experiments use symmetric 0--1 prediction utility.
Policy rankings may change under asymmetric false-positive and
false-negative costs, particularly on the imbalanced streams.

Six limitations concern the design of the comparison. (7) Only the follow-ups
E1--E3 have a commit record showing that their specification preceded their
runs. For the original hypotheses, the order rests on the authors' records
(Section~\ref{sec:design}). (8) The single-arm policies are not one-factor
removals from Full RAR, and the alert-arm removal was not run, so the alert
arm is not isolated. (9) H5 is tested only at the policy level. The matched
per-deployment test was not run because its registered definition is
undefined for events near stream ends. (10) The second learner covers five
streams and four policies, and one FEASIBLE result reverses under it. (11) The 30 seeds vary only canary routing, while 12 of the 13 streams come from five
base datasets; the grouped rank analysis (Table~\ref{tab:lobo}) addresses the second point, not the first. (12) Finally, the payloads record neither the
times at which alerts were charged nor per-stage routed counts. Whether
alerts repeat, and whether every commit completed both canary stages,
therefore cannot be audited from them after the fact. Commit is reachable
only from the second canary stage once its labeled window has filled. The
loop enforces this transition, and the harness's state-transition tests
check it: they cover a full Serving--Canary~1--Canary~2--Serving cycle and
rollback on a failure in either stage. The historical runs, however, carry
no per-stage trace of it.

\subsection{What the Matrix Taught Us About the Instrument}
Four further limitations were discovered by running the study rather than
designing it, and belong beside those above. \emph{(13) The ratified cost
grid was mis-centered.} $\rho \in \{0.1, 0.5, 1, 5\}$ was chosen a priori
around the normalization $C_e = 1$. Across the whole grid, however, the cost
term moves net utility by well under one percent; every measured operating
window opens one to two orders of magnitude higher. We kept the grid as
specified and reported the full break-even curve. A production calibration
is needed to determine whether realistic cost ratios fall inside these
observed windows. \emph{(14) Estimator precision is not decision quality.}
The stratified canary estimator is unbiased and 2.3$\times$ tighter than
i.i.d.\ routing, yet on ELEC2 it does not yield higher utility. A more precise estimate from a 300-instance window can produce a more confident decision on evidence that is not representative of the next 10{,}000 instances. Canary
decision quality is bounded by window representativeness, not measurement
precision. This predicts that widening $W_2$ helps more than tightening the
estimator, which the specified $(W_1, W_2)$ cells test.

\emph{(15) Parameter effects are stream properties.} Ten of twenty-one
one-factor cells change sign between stream groups, and four do so while
resolved on both sides. A pooled sensitivity table is a statement about an
average that may be false of every member. Our defaults are therefore
defaults for the streams we ran. An adopter should re-run the one-factor
grid on their own stream before trusting any of them. \emph{(16) The Gate-2
sample penalty is real and the guard gap is fixed.} Full RAR trains on
$\approx \theta_L W_{\text{train}}$ labeled rows by construction. That is
the right trade under real drift and the wrong one under corruption without
concept change, as Airlines shows. The fixed guard $g$ cannot track a
detection lag that varies by two orders of magnitude across streams. It
removes bias on abrupt drift, leaves most of it on recurrent and corrupted
streams, and adds a little on gradual drift.

\section{Ethical Considerations}
Automated rollback in high-stakes domains (credit, healthcare, fraud) can
silently restore a model whose known deficiencies motivated the retrain.
RAR's alert-hold state partially mitigates this by requiring a monitored hold, with alerts, after every rollback, but deployment in such domains should keep a human in
the escalation path. Automated retraining can also entrench feedback loops
and disparate subgroup impact if drift response is triggered by aggregate
metrics alone. Extending the gates with subgroup-level conditions is
important future work. All experiments use public or synthetic data; no
proprietary or personal data is used.

\section{Reproducibility Record}
\label{sec:reprorecord}
\emph{What ships.} The artifact contains the pipeline, manifests, and checksums; it does not redistribute third-party data. Provider, citation, and license status are disclosed for each dataset. Every cell is traceable to a seed, to a canonical
configuration hash computed identically from either entry point, and to a
PASS/FAIL/MISSING audit record. Every follow-up cell also carries its
producing commit. All \runtworaw{} base-matrix payloads ship with the
artifact, as do the cost-aware payloads, the H6 payloads, and the E1, E2a
and E3 follow-up payloads. Every table in Section~\ref{sec:results} is
regenerated by script, either from these payloads or from campaign
summaries derived from them. The base-matrix, cost-aware, H6 and follow-up
tables were rechecked against the raw cells. The current API reproduces the
base matrix's Full RAR cells exactly (\apiequiv{}/\apiequivtotal{}).

\emph{What was verified independently.} Raw downloads are verified byte for
byte, and prepared streams by the checksum of their decompressed content.
Compressed-file checksums are provenance only, because compressed bytes
depend on the interpreter's zlib build. A clean clone of the release commit
on a second machine, following only the published pipeline, reproduced all
nine raw inputs byte for byte and all eight prepared real streams by
content. The six semi-synthetic streams' content checksums are verified
against the archived files. The post-hoc H6 follow-up stream is regenerated
from its released configuration, and the build checks its own geometry, but
no checksum was recorded for it. A second machine received all 8,148 result files and
verified each against its manifest checksum. It also recomputed the H6
rollback audit, with identical results. The release archive is tested by
extracting it outside the repository and running the test suite there.

\emph{What is missing.} The producing commit of the base matrix was not
stamped into its payloads, and that checkout is no longer accessible. The
run is therefore identified by API version, canonical configuration hash,
cell identities and retained payloads. Commit stamping was added afterward.

\emph{What the process caught.} The decisions log in the artifact records
every defect found during the study, including those that invalidated
earlier results. Here we summarize the classes that could have changed a
reported number. On three occasions, sweep files enumerated cells the
specified protocol did not contain while every value-level check passed.
This would have put twelve unspecified cells into the ablation table as single-factor
rows. Once, a ratified value reached only one of two entry points, so
demonstration runs measured a configuration different from the frozen one even though the freeze check passed. Our own reporting path failed three times: running a generator out of order, a version-specific string construct that
dropped cells from a count, and a diagnostic contract enforced in memory but
not on disk. And artifact checksums were at one point verified only on the
machine that wrote them, which is why prepared streams are now verified by
content. Three rules followed: a sweep file's cell count is a ratified
quantity; a ratified value is enforced everywhere it can be supplied, not
everywhere it happens to be defined; and verification output is not
self-interpreting and must be read against a specified expectation. In our
experience, these rules
are the part of a benchmark that reviewers cannot see and adopters most
need.

An archival DOI has been reserved and will be made public on acceptance. Until then, the
artifact is available to the editor and reviewers on request. Section~\ref{sec:hypotheses}
notes briefly how large language models were used; the full disclosure is
under Statements and Declarations, after the Conclusion.

\section{Conclusion}
We introduced RAR, a three-arm decision framework for drift response in
production ML. It unifies retraining, monitored escalation, and
canary-gated rollback in a single costed state machine. We evaluated it
under a prospectively specified protocol across thirteen streams, seven
policies, and a broad range of retraining costs, with analytic break-even
curves for the cost-blind policies. The headline finding is a map rather
than a winner: no policy dominates across streams. The three-arm policy pays
only inside a bounded window of retraining cost, and the occurrence of that
window is associated with the framework's latency-feasibility condition. What the three-arm policy adds beyond mean utility is operational. It incurs a fraction (one-eighth to one-third) of the pre-onset retrains of detector-triggered retraining. On the post-hoc follow-up stream, it resumes adaptation after a corruption-triggered hold where a single-tier cooldown stays locked.
H2 was not supported on its specified instrument, and H3 held only in its
false-retrain component. Policy-level evidence ran against H5. Adding the
canary to Binary lowered net utility on most streams, while removing the
canary---and consequently the rollback arm---from RAR raised net utility on
11 of 13 streams at $\rho = 1$. H6 was null on its specified stream. All
four are reported with their causes. Beyond the specific results, the
paper's larger claim is methodological: \emph{what to do about drift} is a
decision problem worthy of the same rigor the community has long applied to
\emph{whether drift occurred}. The claim has a second part: a benchmark that
specifies its hypotheses in
advance, freezes its instrument, and audits its own defects can say
precisely where a policy helps and where it does not.

\backmatter

\bmhead{Acknowledgements}
We thank the maintainers of the USP Data Stream Repository, the UCI Machine
Learning Repository, OpenML, and the River library for the public streams and
tooling on which this study depends. The views expressed are the authors'
own and do not represent the positions of their employers.

\section*{Statements and Declarations}

\bmhead{Funding}
The authors did not receive support from any organization for the submitted
work.

\bmhead{Competing interests}
The authors have no relevant financial or non-financial interests to
disclose.

\bmhead{Ethics approval and consent to participate}
Not applicable. The study uses public benchmark data streams and
semi-synthetic streams derived from them; no human participants, personal
data, or animals are involved.

\bmhead{Consent for publication}
Not applicable.

\bmhead{Data availability}
All real data streams are public and are obtained from their providers: the
INSECTS and NOAA Weather streams from the USP Data Stream Repository
\citep{souza2020challenges}, Electricity through the River library
\citep{montiel2021river}, Forest Covertype and Poker Hand from the UCI Machine
Learning Repository, and Airlines from OpenML. No third-party data are
redistributed; the released pipeline downloads each stream from its provider,
verifies the downloaded files byte for byte, and verifies each prepared
stream by the SHA-256 checksum of its decompressed content. Compressed-file
checksums are recorded as provenance but are not a reproduction criterion,
because the compressed bytes depend on the zlib library linked into the
Python interpreter; a rebuilt stream should match in content and may differ
in compressed bytes. The semi-synthetic streams (Section~\ref{sec:datasets};
the corruption stream in both its modes) and the H6 follow-up stream are
deterministic functions of the real streams and are regenerated by the
released injector (seed 007). The per-cell result payloads for every
campaign reported here, the configurations, the dataset cards, and the
decisions log are archived on Zenodo under the reserved DOI
10.5281/zenodo.22829389, which will be made public on acceptance; they are
available to the editor and reviewers on request before then.

\bmhead{Code availability}
The experimental harness, the analysis and table-generation scripts, and
the data acquisition pipeline are archived with the data under the same DOI
and are available to the editor and reviewers on request before
publication.

\bmhead{Author contributions}
A.S. (Ali Sadeghian) ran the experimental campaigns and follow-up
experiments, performed the evaluation and audits, and verified the reported
results. A.Sh. (Alireza Shafaei) built and maintained the experimental
harness, the data acquisition and provenance pipeline, the reproduction
tooling, and the archival artifact. A.S. and A.Sh. contributed equally. H.H. conceived the study, designed the
framework and the experimental protocol, supervised the work, and wrote the
manuscript. B.H.K. supervised the students' research at Sharif and reviewed the manuscript. All authors reviewed and approved the final manuscript.

\bmhead{Use of generative AI}
Large language models (Anthropic Claude and OpenAI ChatGPT) were used to
draft and edit parts of the manuscript text, to assist with writing and
debugging the experimental harness and its tests, to assist with the
editorial drafting of decisions-log entries documenting protocol choices the
authors made, to write table-generation and verification scripts, and to
review drafts. The experimental design, hypotheses, protocol decisions,
analyses, interpretations, and experiments were determined by the authors.
Every reported number is generated by the released scripts from the
per-cell payloads or from campaign summaries derived from them. The authors
reviewed all AI-assisted text and code and take full responsibility for the
submitted work.

\begin{appendices}

\section{Glossary}
\label{app:glossary}
This appendix collects, in one place, every symbol and term of art used in
the paper. The first table below covers the stream model, cost model, and
objective; the second covers gates, windows, the canary protocol, and the
hold states. Defaults are the values used in the main experiments; every
parameter listed with a default is exposed in the released configuration
files and swept in the ablations of Section~\ref{sec:ablations}.

\subsection{Notation}

\noindent\textbf{Stream model, costs, and objective.}\par\vspace{4pt}
\noindent
\footnotesize
\setlength{\tabcolsep}{4pt}
\begin{tabular}{@{}l p{0.72\columnwidth}@{}}
\toprule
Symbol & Meaning \\
\midrule
$S = ((x_t, y_t))_{t=1}^{T}$ & Data stream of feature--label pairs; $T$ is the stream length. \\
$x_t,\, y_t,\, \hat{y}_t$ & Features, true label, and the \emph{served} prediction at time $t$. \\
$P_t(X, Y)$ & Time-varying joint distribution generating the stream. \\
$\tau$ & Label delay: $y_t$ becomes visible at time $t + \tau$. \\
$\ell_t(W)$ & Label availability --- fraction of window $W$ labeled by time $t$ (Eq.~\ref{eq:labelavail}). \\
$\pi,\, a_t$ & Drift-response policy; the action it selects at $t$ (Eq.~\ref{eq:actions}). \\
$M_{\text{old}},\, M_{\text{new}}$ & Incumbent model; retrained challenger under canary evaluation. \\
$u(\hat{y}_t, y_t)$ & Per-prediction utility; $1$ if correct, $0$ otherwise (binary default). \\
$U(\pi; S)$ & Net utility of policy $\pi$ on stream $S$ (Eq.~\ref{eq:netutility}). \\
$C_e$ & Cost of one prediction error; normalized to $C_e = 1$. \\
$C_r,\, C_a,\, C_b$ & Cost of one retrain, alert, rollback. Defaults: $C_a = 0.05\,C_r$, $C_b = 0.2\,C_r$. \\
$\rho = C_r / C_e$ & Cost ratio; regimes $\rho \in \{0.1, 0.5, 1, 5\}$. \\
$N_r,\, N_a,\, N_b$ & Number of retrain, alert, and rollback events over the stream. \\
\bottomrule
\end{tabular}
\normalsize

\vspace{10pt}
\noindent\textbf{Gates, windows, canary, and hold states.}\par\vspace{4pt}
\noindent
\footnotesize
\setlength{\tabcolsep}{4pt}
\begin{tabular}{@{}l p{0.70\columnwidth}@{}}
\toprule
Symbol & Meaning \\
\midrule
$D$ & Pluggable drift detector (ADWIN, DDM, KSWIN, Page--Hinkley, ECDD). \\
$\hat{t}_d$ & Detector-estimated drift onset. \\
$\theta_L$ & Label-availability threshold for Gate~2 (default $0.7$). \\
$\theta_R$ & Minimum incumbent degradation for Gate~3. \\
$\hat{\Delta}_t$ & Degradation estimate $\hat{A}_{\text{pre}} - \hat{A}_{\text{cur}}$ (Eq.~\ref{eq:benefit}). \\
$\hat{A}_{\text{pre}},\, \hat{A}_{\text{cur}}$ & Incumbent rolling accuracy over the reference and current windows. \\
$\hat{A}_{\text{old}},\, \hat{A}_{\text{new}}$ & Incumbent and challenger accuracy on the routed canary sample. \\
$g$ & Guard gap: the reference window ends at $\hat{t}_d - g$ to exclude instances contaminated by detection latency. \\
$W_{\text{train}}$ & Retraining window, anchored at $\hat{t}_d$, extending to $t - \tau$. Feasibility: $W_{\text{train}} \geq \tau / (1 - \theta_L)$. \\
$W_{\text{ref}},\, W_{\text{cur}}$ & Reference and current windows for the Gate-3 comparison. \\
$t^{*}$ & Earliest admissible retrain: $\hat{t}_d + \tau + \lceil \theta_L W_{\text{train}} \rceil - 1$. \\
$p_1$ & Fraction of traffic routed to $M_{\text{new}}$ in Stage~1 (default $0.2$). \\
$W_1,\, W_2$ & Canary stage windows, counted in \emph{labeled} routed instances. \\
$\delta$ & Rollback margin: R1 fires if $\hat{A}_{\text{new}} < \hat{A}_{\text{old}} - \delta$. \\
$q_t$ & Feature-quality statistic driving the pre-filter and condition R2. \\
$\kappa_{\text{canary}}$ & Wall-clock timeout per canary stage; expiry fails closed. \\
$m_{\text{clear}}$ & Consecutive clean $q_t$ assessments that release a hard hold early (default 200). \\
$\kappa_{\text{short}},\, \kappa_{\text{long}}$ & Cooldowns for the soft (quality) and hard (corruption) hold states. \\
$\kappa_{\min}$ & Minimum evidence window before the emergency override may fire. \\
$\Delta_{\text{crit}}$ & Override severity threshold; requires $\Delta_{\text{crit}} \gg \theta_R$. \\
$W_{\text{mon}}$ & Secondary monitoring window refreshed by each alert. \\
\bottomrule
\end{tabular}
\normalsize
\vspace{6pt}

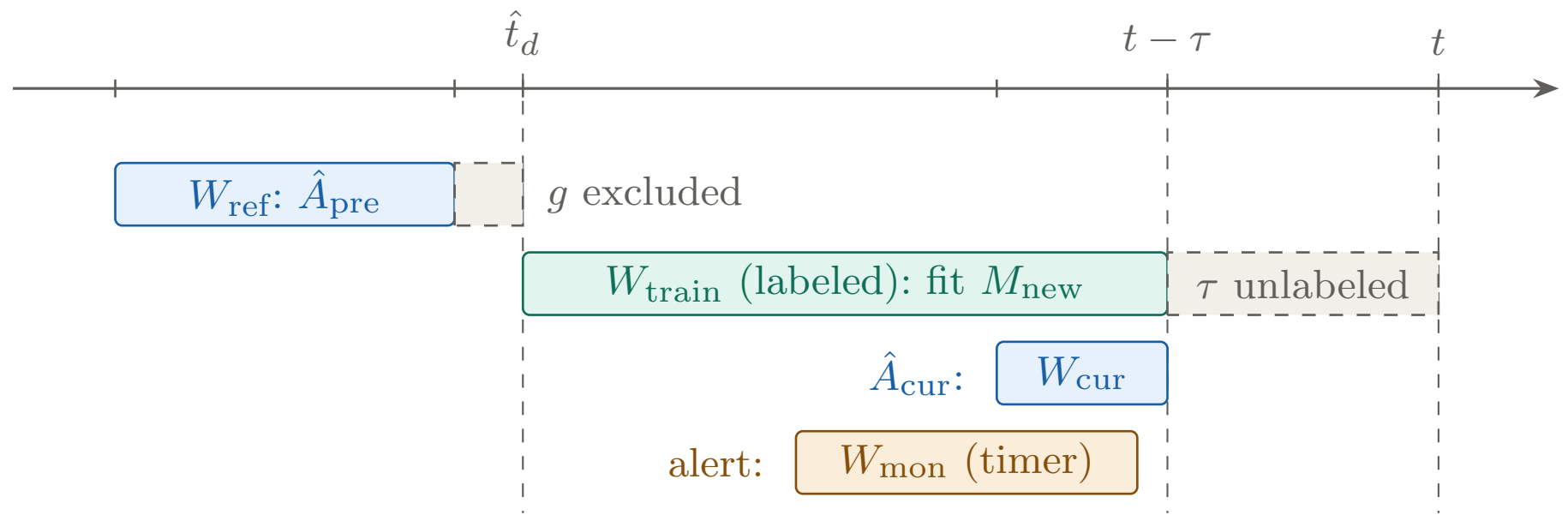
\begin{figure}[h]
  \centering
  \begin{tikzpicture}[font=\scriptsize, x=1.55cm, y=1.15cm]
    \draw[-{Stealth[length=2mm]}, line width=0.6pt, draw=bgray]
      (0,0) -- (7.7,0);
    \foreach \x in {0.51,2.20,2.54,4.90,5.75,7.10}
      \draw[bgray, line width=0.5pt] (\x,0.06) -- (\x,-0.06);
    \draw[bgray, dashed, line width=0.4pt] (2.54,0.10) -- (2.54,-2.85);
    \draw[bgray, dashed, line width=0.4pt] (5.75,0.10) -- (5.75,-2.85);
    \draw[bgray, dashed, line width=0.4pt] (7.10,0.10) -- (7.10,-2.85);
    \node[anchor=south, text=bgray] at (2.54,0.12) {$\hat{t}_d$};
    \node[anchor=south, text=bgray] at (5.75,0.12) {$t-\tau$};
    \node[anchor=south, text=bgray] at (7.10,0.12) {$t$};

    \draw[fill=stblue, draw=bblue, line width=0.5pt, rounded corners=1.5pt]
      (0.51,-0.92) rectangle (2.20,-0.50);
    \node[text=bblue] at (1.355,-0.71) {$W_{\text{ref}}$: $\hat{A}_{\text{pre}}$};
    \draw[fill=stgray, draw=bgray, dashed, line width=0.5pt]
      (2.20,-0.92) rectangle (2.54,-0.50);
    \node[anchor=west, text=bgray] at (2.58,-0.71) {$g$ excluded};

    \draw[fill=stteal, draw=bteal, line width=0.5pt, rounded corners=1.5pt]
      (2.54,-1.52) rectangle (5.75,-1.10);
    \node[text=bteal] at (4.145,-1.31) {$W_{\text{train}}$ (labeled): fit $M_{\text{new}}$};
    \draw[fill=stgray, draw=bgray, dashed, line width=0.5pt]
      (5.75,-1.52) rectangle (7.10,-1.10);
    \node[text=bgray] at (6.425,-1.31) {$\tau$ unlabeled};

    \draw[fill=stblue, draw=bblue, line width=0.5pt, rounded corners=1.5pt]
      (4.90,-2.12) rectangle (5.75,-1.70);
    \node[text=bblue] at (5.325,-1.91) {$W_{\text{cur}}$};
    \node[anchor=east, text=bblue] at (4.82,-1.91) {$\hat{A}_{\text{cur}}$:};

    \draw[fill=stamber, draw=bamber, line width=0.5pt, rounded corners=1.5pt]
      (3.90,-2.72) rectangle (5.60,-2.30);
    \node[text=bamber] at (4.75,-2.51) {$W_{\text{mon}}$ (timer)};
    \node[anchor=east, text=bamber] at (3.82,-2.51) {alert:};
  \end{tikzpicture}
  \caption{Window anatomy on the worked example of
  Section~\ref{sec:framework}: $\hat{t}_d = 9{,}500$, $\tau = 800$,
  $g = 200$, evaluated at $t = 12{,}189$. $W_{\text{ref}}$ (1{,}000
  instances, ending at $\hat{t}_d - g$) and $W_{\text{cur}}$ (the newest
  500 \emph{labeled} instances) are frontier-clocked measurement windows;
  $W_{\text{train}}$ (anchored at $\hat{t}_d$, 1{,}890 labeled of 2{,}690)
  is the learning window; $W_{\text{mon}}$ is a live-edge vigilance timer
  started by an alert at $t = 10{,}300$. Dashed gray spans are excluded
  from, or not yet available to, label-dependent computation.}
  \label{fig:windows}
\end{figure}

\subsection{Terms}

\begin{description}
\item[Alert arm] The graduated response fired when drift evidence exists but
the retrain gates fail (conditions A1--A3, Section~4.2): log an escalation,
charge $C_a$, keep serving $M_{\text{old}}$, and escalate to Retrain if
conditions improve. Distinct from \textsc{Continue} by its trigger, cost, and
monitoring side effect.
\item[\textsc{Alert-Hold-Hard}] The post-rollback state entered after a
corruption-triggered (R2) rollback. Retraining is inadmissible for
$\kappa_{\text{long}}$ with \emph{no} override; exits early the moment $q_t$
confirms the corruption has cleared.
\item[\textsc{Alert-Hold-Soft}] The post-rollback state entered after a
quality-triggered (R1/R3) rollback. Retraining is inadmissible for
$\kappa_{\text{short}}$ unless the emergency override fires.
\item[Arm] One of the three non-trivial responses available to a RAR policy:
Retrain, Alert, or Rollback (plus the implicit \textsc{Continue}).
\item[Binary policy] The literature-standard baseline: detector-triggered
retraining with immediate full deployment; no alert arm, no rollback arm.
\item[Blatant corruption] Corruption producing degenerate feature marginals
(e.g., uniform noise) that the upstream pre-filter is designed to catch and
quarantine. One of the two modes of the Semi-Corrupt stream.
\item[Canary (two-stage)] The staged deployment every retrained model must
pass: Stage~1 routes $p_1$ of traffic to $M_{\text{new}}$; Stage~2 gives it
all traffic while $M_{\text{old}}$ is shadow-scored; failure at either stage
triggers rollback (Section~4.3).
\item[Commit] The transition that promotes $M_{\text{new}}$ to incumbent
after both canary stages pass.
\item[Commit bar] The one-sided promotion evidence threshold:
the discordant-pair statistic $z = (n_{01}-n_{10})/\sqrt{n_{01}+n_{10}}$
must reach $z_{\text{commit}}$; the dead band between the rollback margin
and the bar reverts to the incumbent.
\item[Concept drift] A change in $P_t(Y \mid X)$ over time (real drift), as
distinct from covariate shift in $P_t(X)$ and label shift in $P_t(Y)$. \emph{Abrupt}:
sudden change at an onset. \emph{Gradual}: the new concept phases in over a
transition window. \emph{Incremental}: the distribution moves through a
sequence of intermediate states. \emph{Recurrent}: previously seen concepts
return.
\item[Cost-Aware Binary] The one cost-reading policy: Binary plus the
benefit gate $\hat{\Delta} \cdot \hat{H} \geq \rho$ with a frozen
horizon estimator; the fair opponent at high $\rho$, run at real
log-spaced $\rho$ points because its utility is not affine.
\item[Corruption pre-filter] The layer upstream of the detectors that
computes $q_t$ per instance and quarantines hard-corrupt instances from
detector updates, preventing detector memory poisoning
(Section~4.3).
\item[CD diagram] The critical-difference diagram of the Dem\v{s}ar
protocol, visualizing Nemenyi post-hoc results: policies ordered by mean
rank, with statistically indistinguishable groups connected.
\item[Data corruption] An upstream fault that alters the observed features,
changing the observed $P_t(X)$ but not the relation the model should learn.
Observationally
similar to drift for a detector; retraining on it is harmful.
\item[Detector memory poisoning] Contamination of a detector's internal
sliding windows by corrupted instances, causing persistent false alarms even
after rollback; the failure mode the pre-filter prevents.
\item[Emergency override] The severity-gated escape from
\textsc{Alert-Hold-Soft}: re-admits retraining only when all four conditions
hold --- $t - t_{\text{rb}} \geq \kappa_{\min}$, accuracy collapse beyond
$\Delta_{\text{crit}}$, labels available ($\ell_t \geq \theta_L$), and clean
$q_t$ (Section~4.4).
\item[Escalation] The Alert-to-Retrain path: an alert episode converts to a
retrain when the blocking gate later passes (e.g., labels accumulate past
$\theta_L$ while the detector flag is still active).
\item[False retrain] A retrain triggered before the true drift onset;
measurable only on streams with ground-truth onsets.
\item[Gate] One of the three jointly required admissibility tests for the
retrain arm: drift confirmed (G1), labels available (G2), incumbent
degraded (G3).
\item[Ground-truth onset] The exact instance index at which drift or
corruption was injected into a semi-synthetic stream, recorded in the
stream's metadata; enables recovery-latency and false-retrain metrics.
\item[Label frontier] The boundary $t - \tau$ behind which labels exist.
Every label-dependent quantity (accuracies, $\ell_t$, canary comparisons)
is clocked at the frontier; feature-only and timer quantities ($q_t$,
cooldowns) run at the live edge $t$ (Section~\ref{sec:formulation}).
\item[Ledger (served-prediction rule)] The accounting rule that the
prediction actually served enters the utility ledger --- including
$M_{\text{new}}$'s canary errors --- with routing fixed by a seeded RNG
before the label is observed; shadow predictions are never ledgered.
\item[Negative control] A stream on which a mechanism should \emph{not}
fire; Poker Hand serves this role for the rollback arm and the pre-filter.
\item[Prequential evaluation] Test-then-train: the model predicts on each
instance before that instance's label is (later) revealed and used for
learning and evaluation.
\item[Operating window] The interval of $\rho$ in which Full RAR
simultaneously beats Binary, Always-Retrain, and No-Response --- bounded below by crossovers against costlier policies and above by crossovers against cheaper ones; reported with per-seed ranges.
\item[Prediction collapse] The label-free signature of a broken
challenger: one emitted class across a window while the incumbent stays
varied on the same instances; the disagreement monitor requires it jointly
with high disagreement before aborting a canary.
\item[Quarantine] Exclusion of a hard-corrupt instance from all detector
updates while it is still served and still ledgered.
\item[Recovery latency] Instances elapsed from the true drift onset until
accuracy returns to its pre-drift level; defined only where ground-truth
onsets exist.
\item[Rollback arm] Reversion to $M_{\text{old}}$ when a retrained model
proves harmful, triggered by canary failure (R1), data corruption (R2), or
sanity failure (R3); charges $C_b$, resets the detector, and enters the
typed hold state.
\item[Rollback--retrain oscillation] The pathology of repeatedly retraining
and rolling back on the same conditions; prevented by the hold cooldowns and
$\kappa_{\min}$.
\item[Sanity baseline] The no-skill (majority-class) predictor;
$M_{\text{new}}$ performing below it on the canary window triggers R3.
\item[Semi-synthetic stream] A real public stream with a seeded, precisely
located disturbance injected (drift or corruption), providing ground truth
that fully real streams lack.
\item[Shadow scoring] Evaluating a model on instances it does not serve
(e.g., $M_{\text{old}}$ during Canary-2), for comparison only.
\item[Step-size check] The $q_t$ test of each feature's mean absolute
successive difference against its clean baseline; it catches resampling
corruption to which marginal mean and variance checks are structurally
blind on near-uniform features.
\item[Subtle corruption] Corruption that evades the pre-filter by
construction (a partial-coverage unit change, sub-threshold nulls), delays
the detector, and poisons the retrain; the specified H2 instrument, run as
follow-up E1 rather than in the base matrix.
\item[Two-tier locking] Typing the post-rollback hold by the rollback's
cause --- corruption (hard, no override) versus quality (soft, override
possible) --- so that oscillation protection does not create blindness to
genuine drift arriving after a rollback (Section~4.4, hypothesis H6).
\end{description}

\section{Canary Window Power Analysis}
\label{app:power}
Both models score the same routed instances, so the canary comparison is
paired and its power depends on the discordant pairs, not on the two
marginal accuracies. Let $\psi = P(\text{models disagree in correctness})$
be the discordance and $\delta$ the true accuracy difference. For a
one-sided McNemar test with critical value $z_{\alpha}$ and power $1-\beta$,
the required number of labeled paired instances is
\citep{connor1987sample}
\begin{equation}
n \;\geq\; \frac{\left(z_{\alpha}\sqrt{\psi} +
z_{\beta}\sqrt{\psi - \delta^2}\right)^2}{\delta^2}.
\end{equation}
Table~\ref{tab:power} evaluates it at the deployed commit bar,
$z_{\alpha} = z_{\text{commit}} = 1$ (one-sided $\alpha \approx 0.16$), and
$1-\beta = 0.8$; at the conventional $\alpha = 0.05$ every required window
roughly doubles.
The frozen $W_2 = 600$ detects $\delta$ between \powdetsixlow{} and
\powdetsixhigh{} over $\psi \in [0.05, 0.20]$, and $W_1 = 300$ between
\powdetthreelow{} and \powdetthreehigh{}; a 1-point difference needs
thousands of instances at any $\psi$. The matrix did not log $\psi$, so we
cannot say where in that range the deployed comparisons sat. The commit bar
is a single look at the end of Stage 2 and is covered by this analysis; R1
is a margin rule applied at interim looks and is not
(Section~\ref{sec:rollbackarm}).

\begin{table}[h]
\caption{Paired (McNemar) power for the canary commit bar at the deployed
$z_{\text{commit}} = 1$ (one-sided $\alpha \approx 0.16$), power 0.8, as a
function of discordance $\psi$.}
\label{tab:power}
\centering\footnotesize
\begin{tabular}{@{}lrrr@{}}
\toprule
 & $\psi = 0.05$ & $\psi = 0.10$ & $\psi = 0.20$ \\
\midrule
\multicolumn{4}{@{}l}{\emph{Required labeled window $n$}} \\
$\delta = 0.05$ & 67 & 135 & 270 \\
$\delta = 0.03$ & 187 & 376 & 753 \\
$\delta = 0.01$ & 1,695 & 3,391 & 6,782 \\
\midrule
\multicolumn{4}{@{}l}{\emph{Detectable $\delta$ at window $n$}} \\
$n = 150$ & 0.033 & 0.047 & 0.067 \\
$n = 300$ & 0.024 & 0.034 & 0.047 \\
$n = 600$ & 0.017 & 0.024 & 0.034 \\
$n = 1,200$ & 0.012 & 0.017 & 0.024 \\
\bottomrule
\end{tabular}

\end{table}

\section{Configuration Space and Run Registry}
The released artifact carries the complete run registry. Every cell of the
base matrix, the cost-aware baseline matrix, the routing cell, the H6
comparison and follow-up, the registered follow-ups E1, E2a and E3, and the
ablation campaign is recorded as a
per-cell JSON payload with its dataset, policy, seed, API version,
canonical configuration hash, all 18 guaranteed metric keys, and the
diagnostic trace; each campaign additionally carries a manifest of planned
cells and a PASS/FAIL/MISSING audit. The 48 ratified configuration values of
API 1.1 and the 49 of API 1.2 are listed in the artifact's frozen-contract
document together with the automated check that verifies a checkout against
them. Every table and figure in Section~\ref{sec:results} is produced by
scripts included in the artifact from these payloads; the scripts and their
required execution order are documented there.

\end{appendices}


\end{document}